\documentclass[times,twocolumn,final]{elsarticle}

\usepackage{prletters}
\usepackage{framed,multirow}

\usepackage{amssymb}
\usepackage{latexsym}
\usepackage{enumitem}

\usepackage{url}
\usepackage{xcolor}
\usepackage[font=small,labelfont=bf]{caption}
\definecolor{newcolor}{rgb}{.8,.349,.1}
\usepackage[normalem]{ulem}
\useunder{\uline}{\ul}{}

\newcommand{\etal}{\textit{et al. }}

\journal{Pattern Recognition Letters}

\begin{document}

\thispagestyle{empty}

\begin{frontmatter}

\title{A Survey of Recent Advances in CNN-based Single Image Crowd Counting and Density Estimation}

\author[1]{Vishwanath A. \snm{Sindagi}\corref{cor1}} 
\cortext[cor1]{Corresponding author:}
\ead{vishwanath.sindagi@rutgers.edu}
\author[2]{Vishal M. \snm{Patel}}

\address[1]{Dept. of Electrical and Computer Engineering, 94 Brett Road, Piscataway, NJ 08854, USA}
\address[2]{Dept. of Electrical and Computer Engineering, 94 Brett Road, Piscataway, NJ 08854, USA}

\begin{abstract}
Estimating count and density maps from crowd images has a wide range of applications such as video surveillance, traffic monitoring, public safety and urban planning. In addition, techniques developed for crowd counting can be applied to related tasks in other fields of study such as cell microscopy, vehicle counting and environmental survey. The task of crowd counting and density map estimation is riddled with many challenges such as occlusions, non-uniform density, intra-scene and inter-scene variations in scale and perspective. Nevertheless, over the last few years, crowd count analysis has evolved from earlier methods that are often limited to small variations in crowd density and scales to the current state-of-the-art methods that have developed the ability to perform successfully on a wide range of scenarios. The success of crowd counting methods in the recent years can be largely attributed to deep learning and publications of challenging datasets. In this paper, we provide a comprehensive survey of recent Convolutional Neural Network (CNN) based approaches that have demonstrated significant improvements over earlier methods that rely largely on hand-crafted representations.  First, we briefly review the pioneering methods that use hand-crafted representations and then we delve in detail into the deep learning-based  approaches and recently published datasets.   Furthermore, we discuss the merits and drawbacks of existing CNN-based approaches and identify promising avenues of research in this rapidly evolving field.
\end{abstract}

\end{frontmatter}


\section{Introduction}
\label{sec:intro}

\begin{figure}[t]
\begin{center}
\begin{minipage}{0.49\linewidth}
\includegraphics[width=\linewidth]{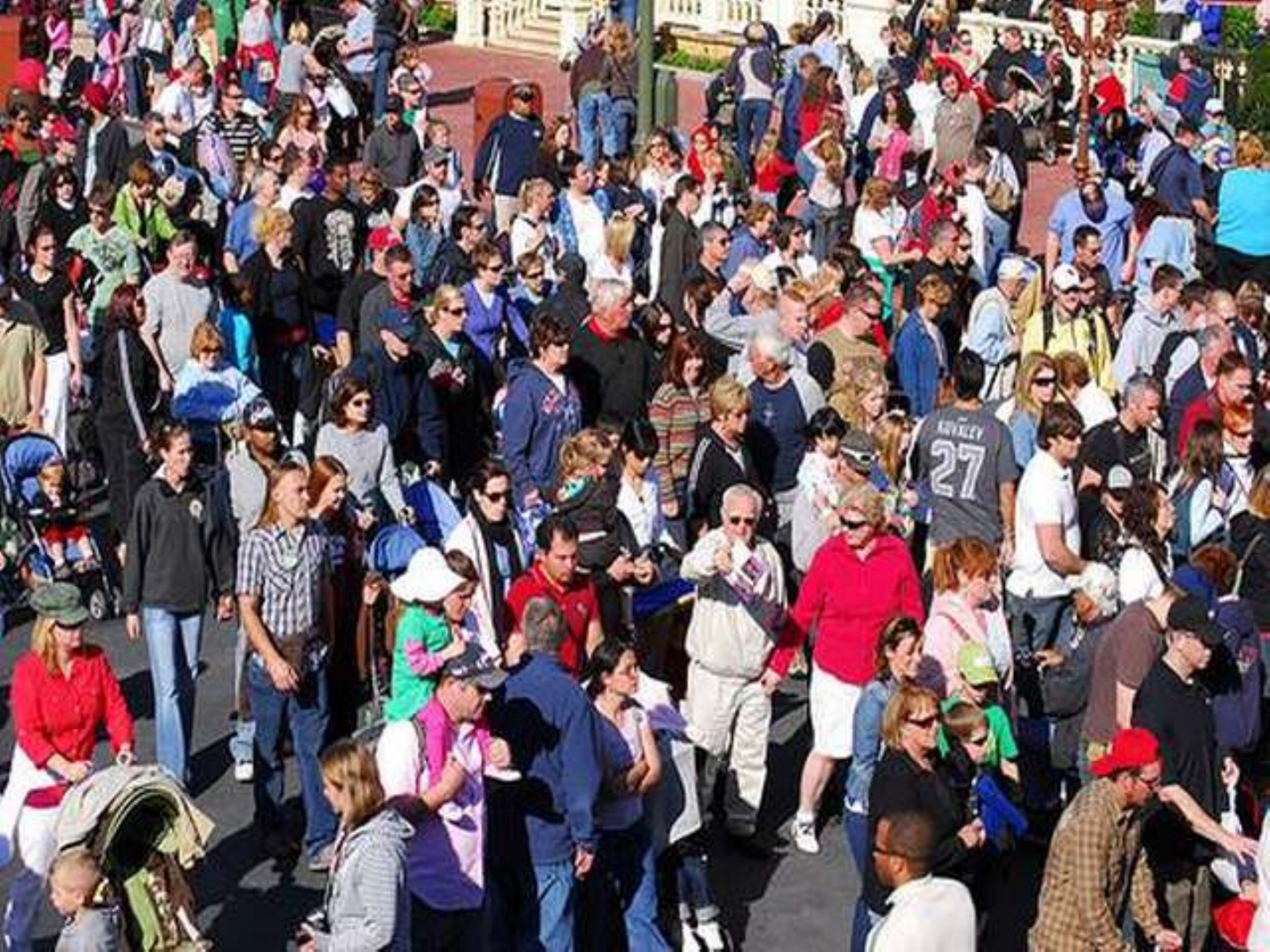}
\vskip-6pt
\captionof*{figure}{(a)}
\end{minipage}%
\hfill
\begin{minipage}{0.49\linewidth}
\includegraphics[width=\linewidth]{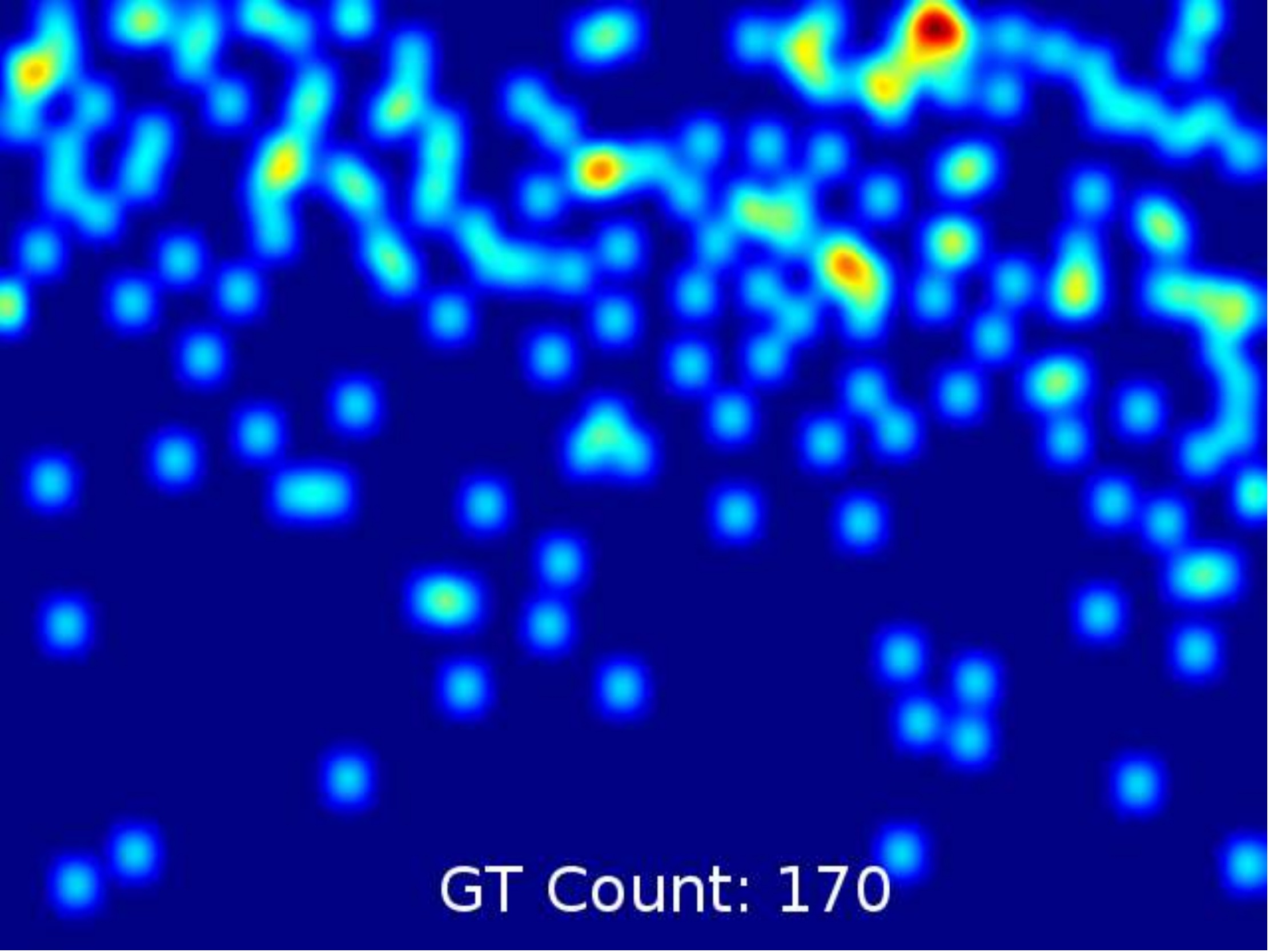}
\vskip-6pt
\captionof*{figure}{(b)}
\end{minipage}
\vskip-10pt
\captionof{figure}{Illustration of density map estimation. (a) Input image (b) Corresponding density map with count.}
\label{fig:task_illustration}
\end{center}

\end{figure}

\begin{figure}[t]
\begin{center}
\begin{minipage}{0.49\linewidth}
\includegraphics[width=\linewidth]{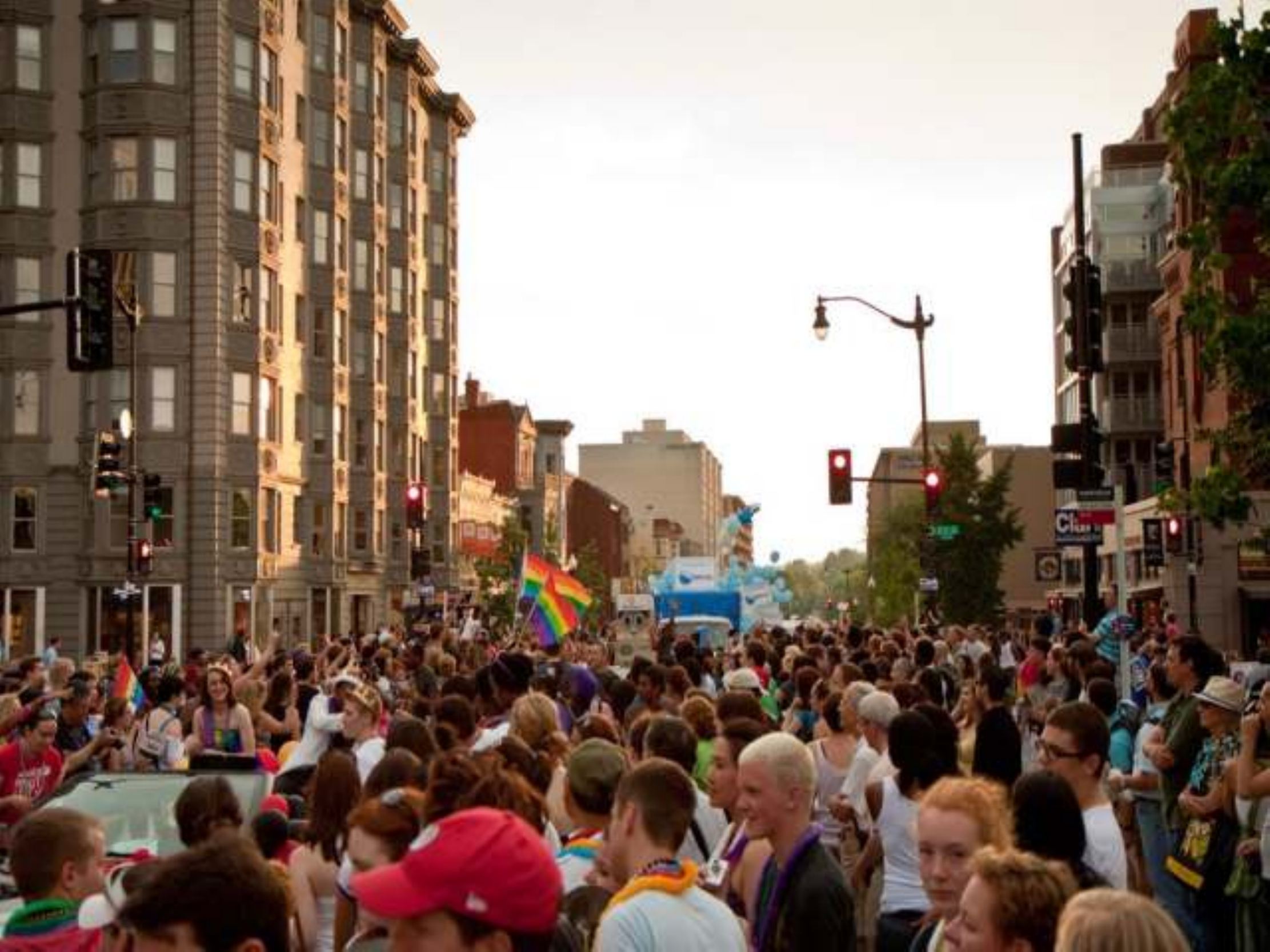}
\vskip-6pt \captionof*{figure}{(a)}
\end{minipage}%
\hfill
\begin{minipage}{0.49\linewidth}
\includegraphics[width=\linewidth]{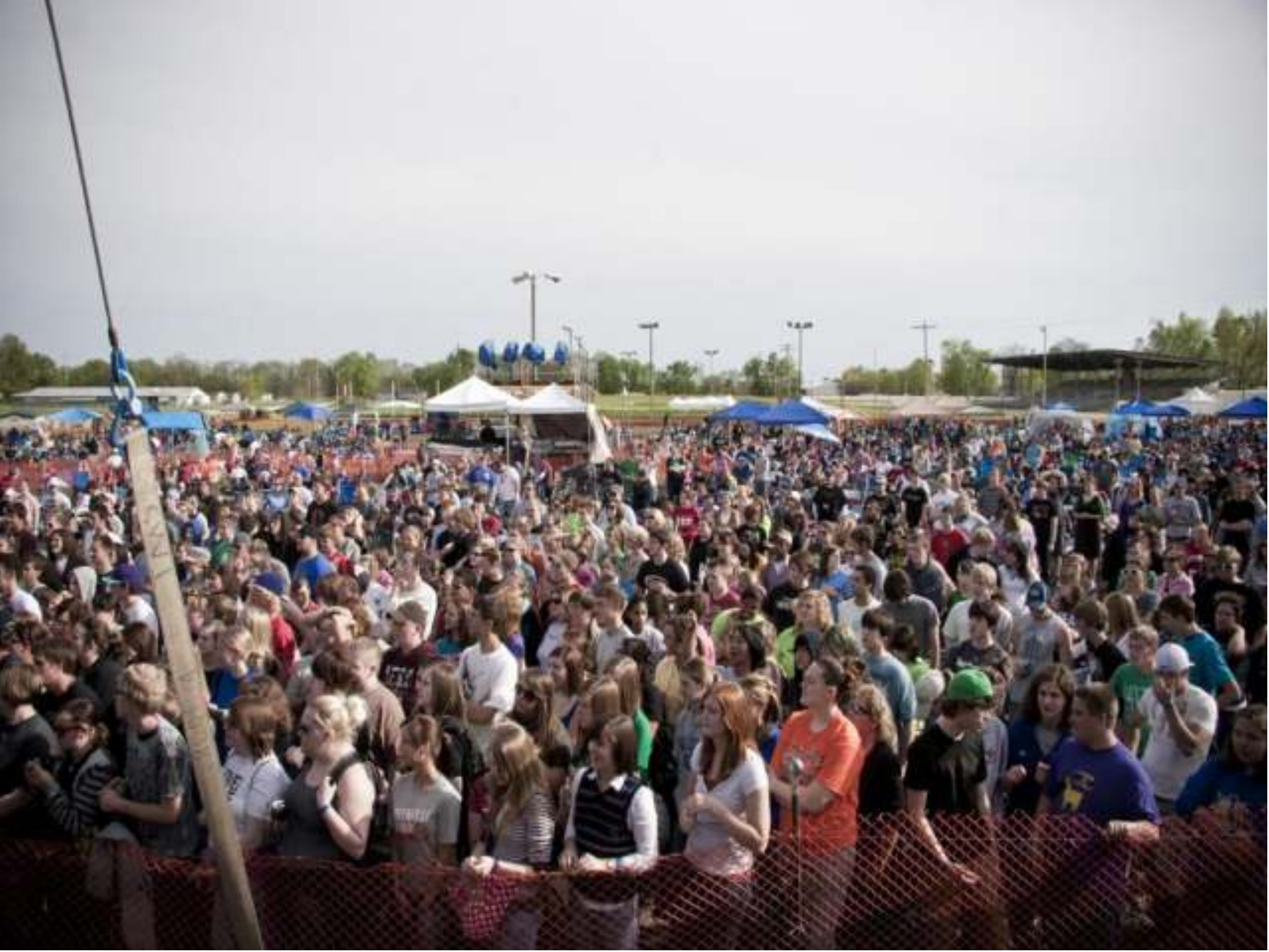}
\vskip-6pt \captionof*{figure}{(b)}
\end{minipage}
\end{center}

\begin{center}
\begin{minipage}{0.49\linewidth}
\includegraphics[width=\linewidth]{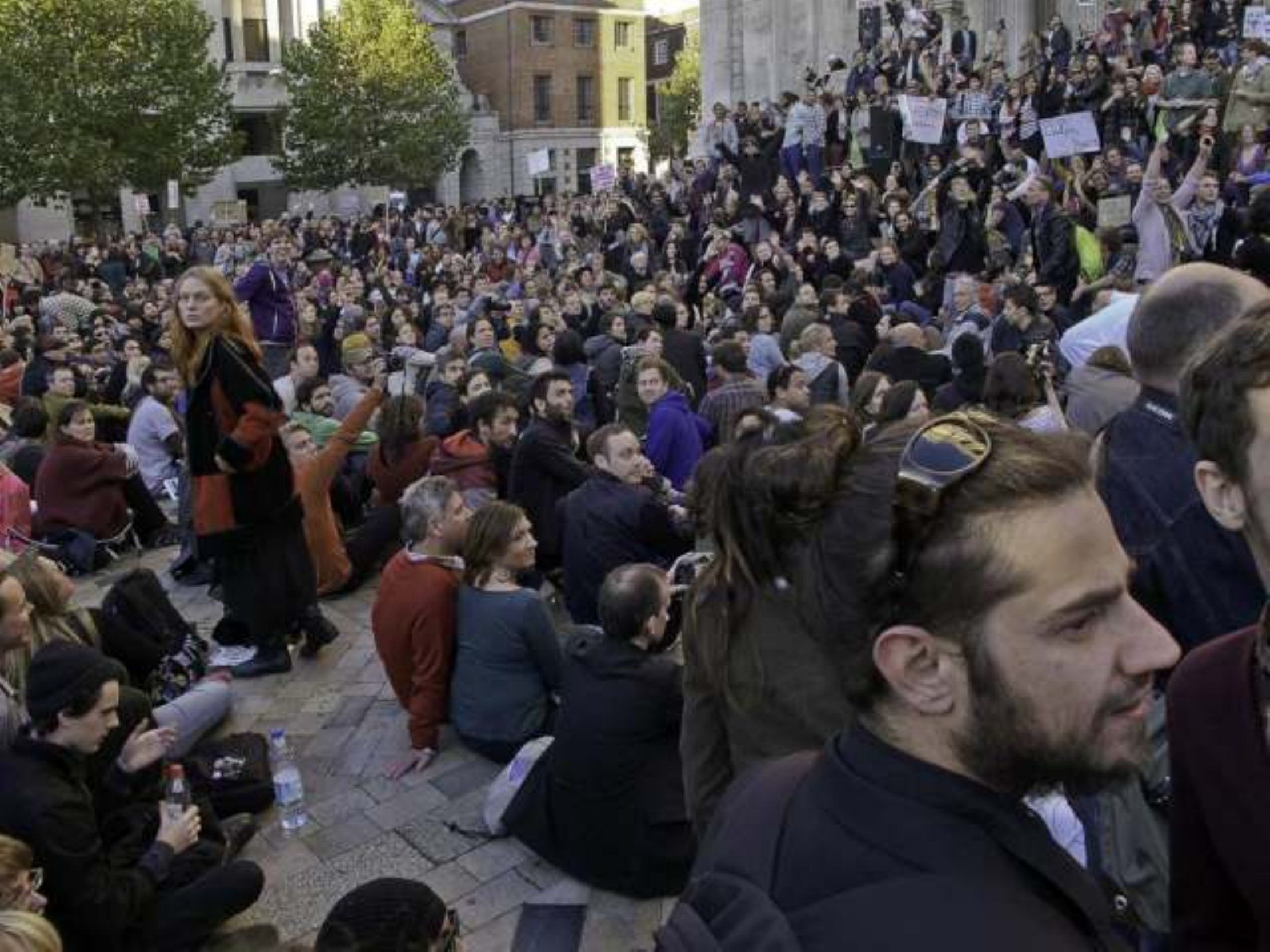}
\vskip-6pt \captionof*{figure}{(c)}
\end{minipage}
\hfill
\begin{minipage}{0.49\linewidth}
\includegraphics[width=\linewidth]{Fig6}
\vskip-6pt \captionof*{figure}{(d)}
\end{minipage}
\vskip-10pt \captionof{figure}{Illustration of various crowded scenes and the associated challenges. (a) Parade (b) Musical concert (c) Public demonstration (d) Sports stadium. High clutter, overlapping of subjects, variation in scale and perspective can be observed across images.}
\label{fig:crowd_scenes}
\end{center}
\end{figure}

Crowd counting aims to count the number of people in a crowded scene where as density estimation aims to map an input crowd image to it's corresponding density map which indicates the number of people per pixel present in the image (as illustrated in Fig. \ref{fig:task_illustration}) and the two problems have been jointly addressed by researchers. The problem of crowd counting and density estimation is of paramount importance  and it is essential for building higher level cognitive abilities  in crowded scenarios such as crowd monitoring \cite{chan2008privacy} and scene understanding \cite{shao2015deeply,zhou2012understanding}. Crowd analysis has attracted significant attention from researchers in the recent past due to a variety of reasons. Exponential growth in the world population and the resulting urbanization has led to an increased number of activities such as sporting events, political rallies, public demonstrations etc. (shown in Fig. \ref{fig:crowd_scenes}), thereby resulting in more frequent crowd gatherings in the recent years. In such scenarios, it is essential to analyze crowd behavior for better management, safety and security. 

Like any other computer vision problem, crowd analysis comes with many challenges such as occlusions, high clutter, non-uniform distribution of people, non-uniform illumination, intra-scene and inter-scene variations in appearance, scale and perspective making the problem extremely difficult. Some of these challenges are illustrated in Fig. \ref{fig:crowd_scenes}. The complexity of the problem together with the wide range of applications for crowd analysis has led to an increased focus by researchers in the recent past. 

Crowd analysis is an inherently inter-disciplinary research topic with researchers from different communities (such as sociology \cite{moussaid2010walking,blumer1951collective}, psychology \cite{aveni1977not}, physics \cite{castellano2009statistical,1971HendersonStatistics}, biology \cite{parrish1999complexity,zhang2010collective}, computer vision and public safety) have addressed the issue from different viewpoints. Crowd analysis has a variety of critical applications of inter-disciplinarian nature:

\noindent \textit{Safety monitoring}: The widespread usage of video surveillance cameras for security and safety purposes in places such as sports stadiums, tourist spots, shopping malls and airports has enabled easier monitoring of crowd in such scenarios. However, traditional surveillance algorithms may break down as they are unable to process high density crowds due to limitations in their design. In such scenarios, we can leverage the results of algorithms specially designed for crowd analysis related tasks such as behavior analysis \cite{saxena2008crowd,ko2008survey}, congestion analysis \cite{zhou2015learning,huang2015congestion}, anomaly detection \cite{li2014anomaly,chaker2017social} and event detection \cite{benabbas2010motion}.

\noindent \textit{Disaster management}: Many scenarios involving crowd gatherings such as sports events, music concerts, public demonstrations and political rallies face the risk of crowd related disasters such as stampedes which can be life threatening. In such cases, crowd analysis can be used as an effective tool for early overcrowding detection and appropriate management of crowd, hence, eventual aversion of any disaster \cite{abdelghany2014modeling,almeida2013crowd}. 

\noindent \textit{Design of public spaces}: Crowd analysis on existing public spots such as airport terminals, train stations, shopping malls and other public buildings \cite{chow2008waiting,sime1995crowd} can reveal important design shortcomings from crowd safety and convenience point of view. These studies can be used for design of public spaces that are optimized for better safety and crowd movement \cite{lu2016study,al2013crowd}. 

\noindent \textit{Intelligence gathering and analysis}: Crowd counting techniques can be used to gather intelligence for further analysis and inference. For instance, in retail sector, crowd counting can be used to gauge people's interest in a product in a store and this information can be used for appropriate product placement \cite{lipton2015video,mongeon2015busyness}.  Similarly, crowd counting can be used to measure queue lengths to optimize staff numbers at different times of the day.  Furthermore, crowd counting can be used to analyze pedestrian flow at signals at different times of the day and this information can be used for optimizing signal-wait times \cite{bernal2014system}. 

\noindent \textit{Virtual environments}: Crowd analysis methods can be used to understand the underlying phenomenon thereby enabling us to establish mathematical models that can provide accurate simulations. These mathematical models can be further used for simulation of crowd phenomena for various applications such as computer games, inserting visual effects in film scenes and designing evacuation plans \cite{gustafson2016mure,perez2016task}.

\noindent \textit{Forensic search}: Crowd analysis can be used to search for suspects and victims in events such as bombing, shooting or accidents in large gatherings.  Traditional face detection and recognition algorithms can be speeded up using crowd analysis techniques which are more adept at handling such scenarios \cite{klontz2013case,barr2014effectiveness}.

These variety of applications has motivated researchers across various fields to develop sophisticated methods for crowd analysis and related tasks such as counting \citep{chan2008privacy,chan2009bayesian,chen2012feature,idrees2013multi,chan2012counting,skaug2016end,ge2009marked,idrees2013multi}, density estimation \citep{lempitsky2010learning,chen2013cumulative,zhang2016single,zhang2015cross,pham2015count,wang2016fast,boominathan2016crowdnet}, segmentation \citep{kang2014fully,dong2007fast}, behaviour analysis \citep{bandini2014towards,shao2014scene,cheng2014recognizing,zhou2012understanding,zhou2015learning,yi2016l0}, tracking  \citep{rodriguez2011density,zhu2014crowd}, scene understanding \citep{shao2015deeply,zhou2012understanding} and anomaly detection \citep{mahadevan2010anomaly,li2014anomaly}. Among these, crowd counting and density estimation are a set of fundamental tasks and they form basic building blocks for various other applications discussed earlier. Additionally, methods developed for crowd counting can be easily extended to counting tasks in other fields such as cell microscopy \cite{wang2016fast,walach2016learning,lempitsky2010learning,chen2012feature}, vehicle counting \cite{onoro2016towards}, environmental survey \cite{french2015convolutional,zhan2008crowd}, etc.

Over the last few years, researchers have attempted to address the issue of crowd counting and density estimation using a variety of approaches such as detection-based counting, clustering-based counting and regression-based counting \cite{loy2013crowd}. The initial work on regression-based methods mainly use hand-crafted features and the more recent works use Convolutional Neural Network (CNN) based approaches. The CNN-based approaches have demonstrated significant improvements over previous hand-crafted feature-based methods, thus, motivating more researchers to explore CNN-based approaches further for related crowd analysis problems. In this paper, we review various single image crowd counting and density estimation methods with a specific focus on recent CNN-based approaches.

Researchers have attempted to provide a comprehensive survey and evaluation of existing techniques for various aspects of crowd analysis \citep{zhan2008crowd,ferryman2014performance,junior2010crowd,li2015crowded,zitouni2016advances}.  Zhan \etal  \cite{zhan2008crowd} and Junior \etal \cite{junior2010crowd} were among the first ones to study and review existing methods for general crowd analysis. Li \etal  \cite{li2015crowded} surveyed different methods for crowded scene analysis  tasks such as crowd motion pattern learning, crowd behavior, activity analysis and anomaly detection in crowds. More recently, Zitouni \etal \cite{zitouni2016advances} evaluated existing methods across different research disciplines by inferring key statistical evidence from existing literature and provided suggestions towards the general aspects of techniques rather than any specific algorithm. While these works focussed on the general aspects of crowd analysis, researchers have studied in detail crowd counting and density estimation methods specifically  \cite{loy2013crowd,saleh2015recent,ryan2015evaluation}. Loy \etal \cite{loy2013crowd} provided a detailed description and comparison of video imagery-based crowd counting and evaluation of different methods using the same protocol. They also analyzed each processing module to identify potential bottlenecks to provide new directions for further research. In another work, Ryan \etal \cite{ryan2015evaluation} presented an evaluation of regression-based methods for crowd counting across multiple datasets and provided a detailed analysis of performance of various hand-crafted features. Recently, Saleh \etal \cite{saleh2015recent} surveyed two main approaches which are direct approach (i.e., object based target detection) and indirect approach (e.g. pixel-based, texture-based, and corner points based analysis).

Though existing surveys analyze various methods for crowd analysis and counting, they however cover only traditional methods that use hand-crafted features and do not take into account the recent advancements driven primarily by CNN-based approaches \cite{shao2015deeply,hu2016dense,zhao2016crossing,boominathan2016crowdnet,skaug2016end,walach2016learning,arteta2016counting,wang2015deep,zhang2016single,zhang2015cross,onoro2016towards,shao2016slicing}
and creation of new challenging crowd datasets \cite{zhang2016data,zhang2015cross,zhang2016single}. While CNN-based approaches have achieved drastically lower error rates, the creation of new datasets has enabled learning of more generalized models. To keep up with the rapidly advancing research in crowd counting, we believe it is necessary to analyze these methods in detail in order to understand the trends.  Hence, in this paper, we provide a survey of recent state-of-the-art CNN-based approaches for crowd counting and density estimation for single images.

Rest of the paper is organized as follows: Section \ref{sec:review_traditional} briefly reviews the traditional crowd counting and density estimation approaches with an emphasis on the most recent methods. This is followed by a detailed survey on CNN-based methods along with a discussion on their merits and drawbacks in Section \ref{sec:survey_cnn}. In Section \ref{sec:datasets_and_results}, recently published challenging datasets for crowd counting are discussed in detail along with results of the state-of-the-art methods. We discuss several promising avenues for achieving further progress in Section \ref{sec:future_research}.  Finally, concluding remarks are made in Section \ref{sec:conclusion}.

\section{Review of traditional approaches}
\label{sec:review_traditional}
Various approaches have been proposed to tackle the problem of crowd counting in images \cite{idrees2013multi,chen2013cumulative,lempitsky2010learning,zhang2015cross,zhang2016single} and videos \cite{brostow2006unsupervised,ge2009marked,rodriguez2011density,chen2015person}. 
Loy \etal \cite{loy2013crowd} broadly classified traditional crowd counting methods based on the approach into the following categories: (1) Detection-based approaches, (2) Regression-based approaches, and (3) Density estimation-based approaches. 

Since the focus of this work is on CNN-based approaches, in this section, we briefly review the detection and regression-based approaches using hand-crafted features for the sake of completeness. In addition, we present a review of the recent traditional methods \cite{idrees2013multi,lempitsky2010learning,pham2015count,wang2016fast,xu2016crowd} that have not been analyzed in earlier surveys. 

\subsection{Detection-based approaches}
Most of the initial research was focussed on detection style framework, where a sliding window detector is used to detect people in the scene \cite{dollar2012pedestrian} and this information is used to count the number of people \cite{li2008estimating}. Detection is usually performed either in the monolithic style or parts-based detection. Monolithic detection approaches \cite{dalal2005histograms,leibe2005pedestrian,tuzel2008pedestrian,enzweiler2009monocular} typically are traditional pedestrian detection methods which train a classifier using features (such as Haar wavelets \cite{viola2004robust}, histogram oriented gradients \cite{dalal2005histograms}, edgelet \cite{wu2005detection} and shapelet \cite{sabzmeydani2007detecting}) extracted from a full body. Various learning approaches such as Support Vector Machines, boosting \cite{viola2005detecting} and random forest \cite{gall2011hough} have been used with varying degree of success. Though successful in low density crowd scenes, these methods are adversely affected by the presence of high density crowds. Researchers have attempted to address this issue by adopting part-based detection methods \cite{felzenszwalb2010object,lin2001estimation,wu2007detection}, where one constructs boosted classifiers for specific body parts such as
the head and shoulder to estimate the people counts in a designated area \cite{li2008estimating}. In another approach using shape learning, Zhao et al. \cite{zhao2008segmentation} modelled humans using 3D shapes composed of ellipsoids, and employed a stochastic process to estimate the number and shape configuration that best explains a given foreground mask in a scene. Ge and Collins \cite{ge2009marked} further extended the idea by using flexible and practical shape models.

\subsection{Regression-based approaches}
Though parts-based and shape-based detectors were used to mitigate the issues of occlusion, these methods were not successful in the presence of extremely dense crowds and
high background clutter. To overcome these issues, researchers attempted to count by regression where they learn a mapping between features extracted from local image patches to their counts  \cite{chan2009bayesian,ryan2009crowd,chen2012feature}. By counting using regression, these methods avoid dependency on learning detectors which is a relatively complex task. These methods have two major components: low-level feature extraction and regression modelling. A variety of features such as foreground features, edge features, texture and gradient features have been used for encoding low-level information. Foreground features are extracted from foreground segments in a video using standard background subtraction techniques. Blob-based holistic features such as area, perimeter, perimeter-area ration, etc. have demonstrated encouraging results \cite{chan2008privacy,chen2012feature,ryan2009crowd}. While these methods capture global properties of the scene, local features such as edges and texture/gradient features such as local binary pattern (LBP), histogram oriented gradients (HOG), gray level co-occurrence matrices (GLCM) have been 
used to further improve the results. Once these global and local features are extracted, different regression techniques such as linear regression \cite{paragios2001mrf}, piecewise linear regression \cite{chan2008privacy}, ridge regression \cite{chen2012feature},  Gaussian process regression and neural network \cite{marana1998efficacy} are used to learn a mapping from low-level feature to the crowd count.

In a recent approach, Idrees \etal \cite{idrees2013multi} identified that no single feature or detection method is reliable enough to provide sufficient information for accurate counting in the presence of high density crowds due to various reasons such as low resolution, severe occlusion, foreshortening and perspective.  Additionally, they observed that there exists a spatial relationship that can be used to constrain the count estimates in neighboring local regions. With these observations in mind, they proposed to extract features using different methods that capture different information. By treating densely packed crowds of individuals as irregular and non-homogeneous texture, they employed Fourier analysis along with head detections and SIFT interest-point based counting in local neighborhoods. The count estimates from this localized multi-scale analysis are then aggregated subject to global consistency constraints. The  three sources, i.e., Fourier, interest points and head detection are then combined with their respective confidences and counts at localized patches are computed independently. These local counts are then globally constrained in a multi-scale Markov Random Field (MRF) framework to get an estimate of count for the entire image.  The authors also introduced an annotated dataset (UCF\textunderscore CC\textunderscore 50) of 50 images containing 64000 humans.

Chen \etal \cite{chen2013cumulative} introduced a novel cumulative attribute concept for learning a regression model when only sparse and imbalanced data are available. Considering that the challenges of inconsistent features along with sparse  and imbalanced (encountered during learning a regression function) are  related, cumulative attribute-based representation for learning a regression model is proposed. Specifically, features extracted from sparse and imbalanced image samples are mapped onto a cumulative attribute space. The method is based on the notion of discriminative attributes used for addressing sparse training data. This method is inherently capable of handling imbalanced data.

\subsection{Density estimation-based approaches}
While the earlier methods were successful in addressing the issues of occlusion and clutter, most of them ignored important spatial information as they were regressing on the global count. In contrast, Lempitsky \etal \cite{lempitsky2010learning} proposed to learn a linear mapping between local patch features and corresponding object density maps, thereby incorporating spatial information in the learning process. In doing so, they avoided the hard task of learning to detect and localize individual object instances by introducing a new approach of estimating image density whose integral over any region in the density map gives the count of objects within that region. The problem of learning density maps is formulated as a minimization of a regularized risk quadratic cost function. A new loss function appropriate for learning density maps is introduced. The entire problem is posed as a convex optimization task which they solve using cutting-plane optimization.

Observing that it is difficult to learn a linear mapping, Pham \etal \cite{pham2015count} proposed to learn a non-linear mapping between  local patch features and density maps. They used random forest regression from multiple image patches to vote for densities of multiple target objects to learn a non-linear mapping. In addition, they tackled the problem of large variation in appearance and shape between crowded image patches and non-crowded ones by proposing a crowdedness prior and they trained two different forests corresponding to this prior.  Furthermore, they were able to successfully speed up the estimation process for real-time performance by proposing an effective forest reduction that uses permutation of decision trees. Apart from achieving real-time performance, another advantage of their method is that it requires relatively less memory to build and store the forest.

Similar to the above approach, Wang and Zou \cite{wang2016fast} identified that though existing methods are effective, they were inefficient from computational complexity point of view. To this effect, they proposed a fast method for density estimation based on subspace learning. Instead of  learning a mapping between dense features and their corresponding density maps, they learned to compute the embedding of each subspace formed by image patches. Essentially, they exploited the relationship between images and their corresponding density maps in the respective feature spaces. The feature space of image patches are clustered and examples of each subspace are collected to learn its embedding. Their assumption that local image patches and their corresponding density maps share similar local geometry enables them to learn locally linear embedding using which  the density map of an image patch can be estimated by preserving the geometry. Since, implementing locally linear embedding (LLE) is time-consuming, they divided the feature spaces  of image patches and their counterpart density maps into subspaces, and computed the embedding of each subspace formed by image patches. The density map of input patch is then estimated by simple classification and mapping with the corresponding embedding matrix. 

In a more recent approach, Xu and Qiu \cite{xu2016crowd} observed that the existing crowd density estimation methods used a smaller set of features thereby limiting their ability to perform better.  Inspired by the ability of high-dimensional features in other domains such as face recognition, they proposed to boost the performances of crowd density estimation by using a much extensive and richer set of features. However, since the regression techniques used by earlier methods (based on Gaussian process
regression or Ridge regression) are computationally complex and are unable to process very high-dimensional features,  they used random forest as the regression model whose tree structure is intrinsically fast and scalable. Unlike traditional approaches to random forest construction, they embedded random projection in the tree nodes to combat the curse of dimensionality and to introduce randomness in the tree construction.

\begin{figure}[t]
\begin{center}
\begin{minipage}{0.95\linewidth}
\includegraphics[width=\linewidth]{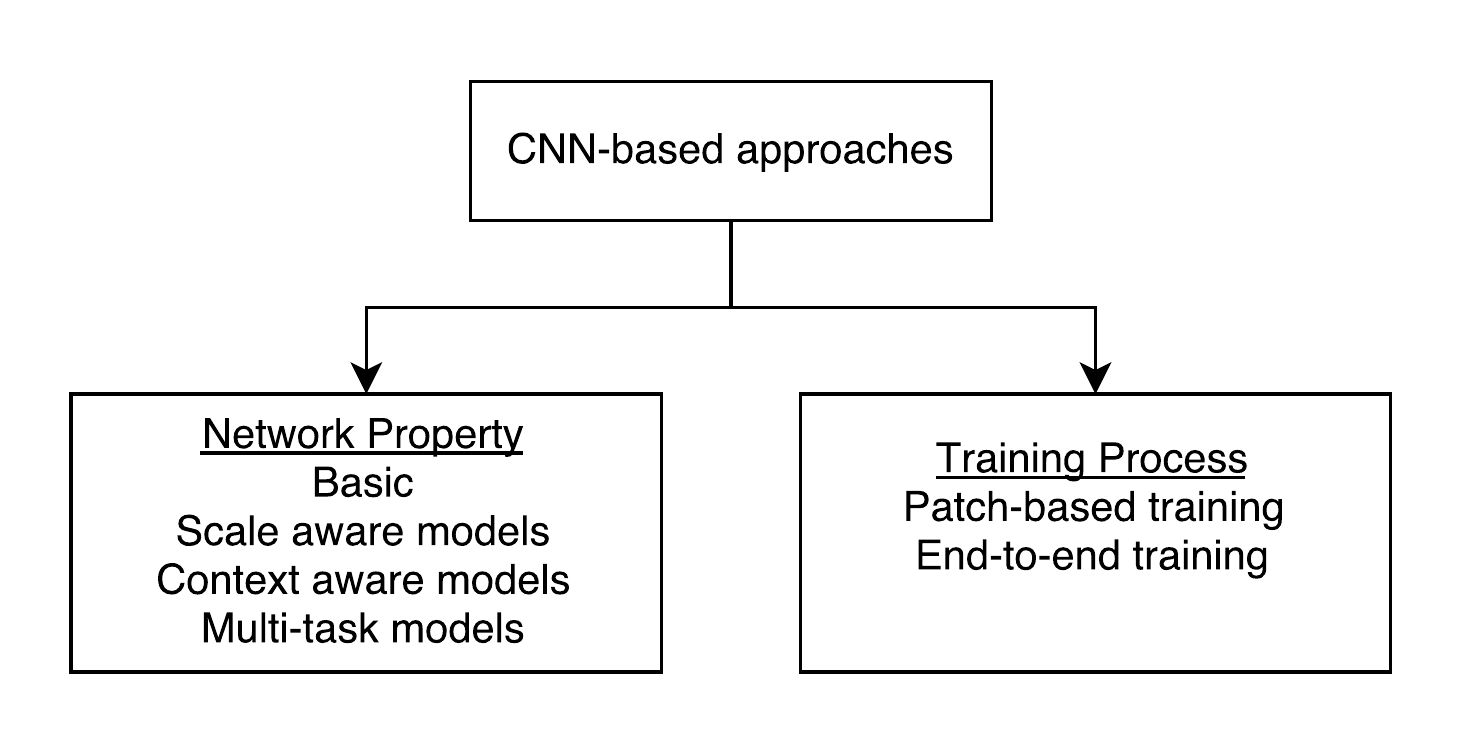}
\end{minipage}%
\vskip-10pt
\captionof{figure}{Categorization of existing CNN-based approaches.}
\label{fig:classification}
\end{center}
\end{figure}

\section{CNN-based methods}
\label{sec:survey_cnn}
The success of CNNs in numerous computer vision tasks has inspired researchers to exploit their abilities for learning non-linear functions from crowd images to their corresponding density maps or corresponding counts. A variety of CNN-based methods have been proposed in the literature. We broadly categorize these methods based on property of the networks  and training approach as shown in Fig. \ref{fig:classification}. Based on the property of the networks, we classify the approaches into the following categories:
\begin{itemize}[noitemsep]
\item \textbf{Basic CNNs}: Approaches that involve basic CNN layers in their networks fall into this category. These methods are amongst initial deep learning approaches for crowd counting and density estimation.
\item \textbf{Scale-aware models}: The basic CNN-based approaches evolved into more sophisticated models that were robust to variations in scale. This robustness is achieved through different techniques such as multi-column or multi-resolution architectures.
\item \textbf{Context-aware models}: Another set of approaches attempted to incorporate local and global contextual information present in the image into the CNN framework for achieving lower estimation errors. 
\item \textbf{Multi-task frameworks}: Motivated by the success of multi-task learning for various computer vision tasks, various approaches have been developed to combine crowd counting and estimation along with other tasks such as foreground-background subtraction and crowd velocity estimation.
\end{itemize}

In an yet another categorization, we classify the CNN-based approaches based on the inference methodology into the following two categories:
\begin{itemize}[noitemsep]
\item \textbf{Patch-based inference}: In this approach, the CNNs are trained using patches cropped from the input images. Different methods use different crop sizes. During the prediction phase, a sliding window is run over the test image and predictions are obtained for each window and finally aggregated to obtain total count in the image. 
\item \textbf{Whole image-based inference}: Methods in this category perform a whole-image based inference. These methods avoid computationally expensive sliding windows. 
\end{itemize}
Table~\ref{tab:classification} presents a categorization of various CNN-based crowd counting methods based on their network property and inference process.

\begin{table}[htp!]
\centering
\caption{Categorization of existing CNN-based approaches.}
\label{tab:classification}
\resizebox{0.99\linewidth}{!}{
\begin{tabular}{|l|l|l|}
\hline
                                                 & \multicolumn{2}{c|}{Category}                                                                                            \\ \hline
Method                                           & \begin{tabular}[c]{@{}l@{}}Network \\ property\end{tabular} & \begin{tabular}[c]{@{}l@{}}Inference\\ process\end{tabular} \\ \hline
Fu \etal \cite{fu2015fast}                       & Basic                                                       & Patch-based                                                \\ \hline
Wang \etal \cite{wang2015deep}                   & Basic                                                       & Patch-based                                                 \\ \hline
Zhang \etal \cite{zhang2015cross}                & Multi-task                                                  & Patch-based                                                \\ \hline
Boominathan \etal \cite{boominathan2016crowdnet} & Scale-aware                                                 & Patch-based                                                \\ \hline
Zhang \etal \cite{zhang2016single}               & Scale-aware                                                 & Whole image-based                                                 \\ \hline
Walach and Wolf \cite{walach2016learning}        & Basic                                                       & Patch-based                                                \\ \hline
Onoro \etal \cite{onoro2016towards}              & Scale-aware                                                 & Patch-based                                                \\ \hline
Shang \etal \cite{skaug2016end}                  & Context-aware                                               & Whole image-based                                                 \\ \hline
Sheng \etal \cite{sheng2016crowd}                & Context-aware                                               & Whole image-based                                                 \\ \hline
Kumagai \etal \cite{kumagai2017mixture}                & Scale-aware                                               & Patch-based                                                 \\ \hline
Marsden \etal \cite{marsden2016fully}                & Scale-aware                                               & Whole image-based                                                 \\ \hline
Mundhenk \etal \cite{mundhenk2016large}          & Basic                                                       & Patch-based                                                \\ \hline
Artetta \etal \cite{arteta2016counting}          & Multi-task                                                  & Patch-based                                                \\ \hline
Zhao \etal \cite{zhao2016crossing}               & Multi-task                                                  & Patch-based                                                \\ \hline
Sindagi \etal \cite{sindagi2017cnnbased}               & Multi-task                                                  & Whole image-based                                                \\ \hline
Sam \etal \cite{sam2017switching}               & Scale-aware & Patch-based                                                \\ \hline
Kang \etal \cite{zhao2016crossing}               & Basic & Patch-based                                                \\ \hline
\end{tabular}
}
\end{table}

\subsection{Survey of CNN-based methods}
In this section, we review various CNN-based crowd counting and density estimation methods along with their merits and drawbacks.

Wang \etal \cite{wang2015deep} and Fu \etal \cite{fu2015fast} were among the first ones to apply CNNs for the task of crowd density estimation. Wang \etal proposed an end-to-end deep CNN regression model for counting people from images in extremely dense crowds. They adopted AlexNet network \cite{krizhevsky2012imagenet} in their architecture where the final fully connected layer of 4096 neurons is replaced with a single neuron layer for predicting the count. Besides, in order to reduce false responses background like buildings and trees in the images, training data is augmented with additional negative samples whose ground truth count is set as zero. In a different approach, Fu \etal proposed to classify the image into one of the five classes: very high density, high density, medium density, low density and very low density instead of estimating density maps. Multi-stage ConvNet from the works of Sermanet \etal \cite{sermanet2012convolutional} was adopted for better shift, scale and distortion invariance. In addition, they used a cascade of two classifiers to achieve boosting in which the first one specifically samples misclassified images whereas the second one reclassifies rejected samples. 

\begin{figure}[htp!]
\begin{center}
\begin{minipage}{1\linewidth}
\includegraphics[width=\linewidth]{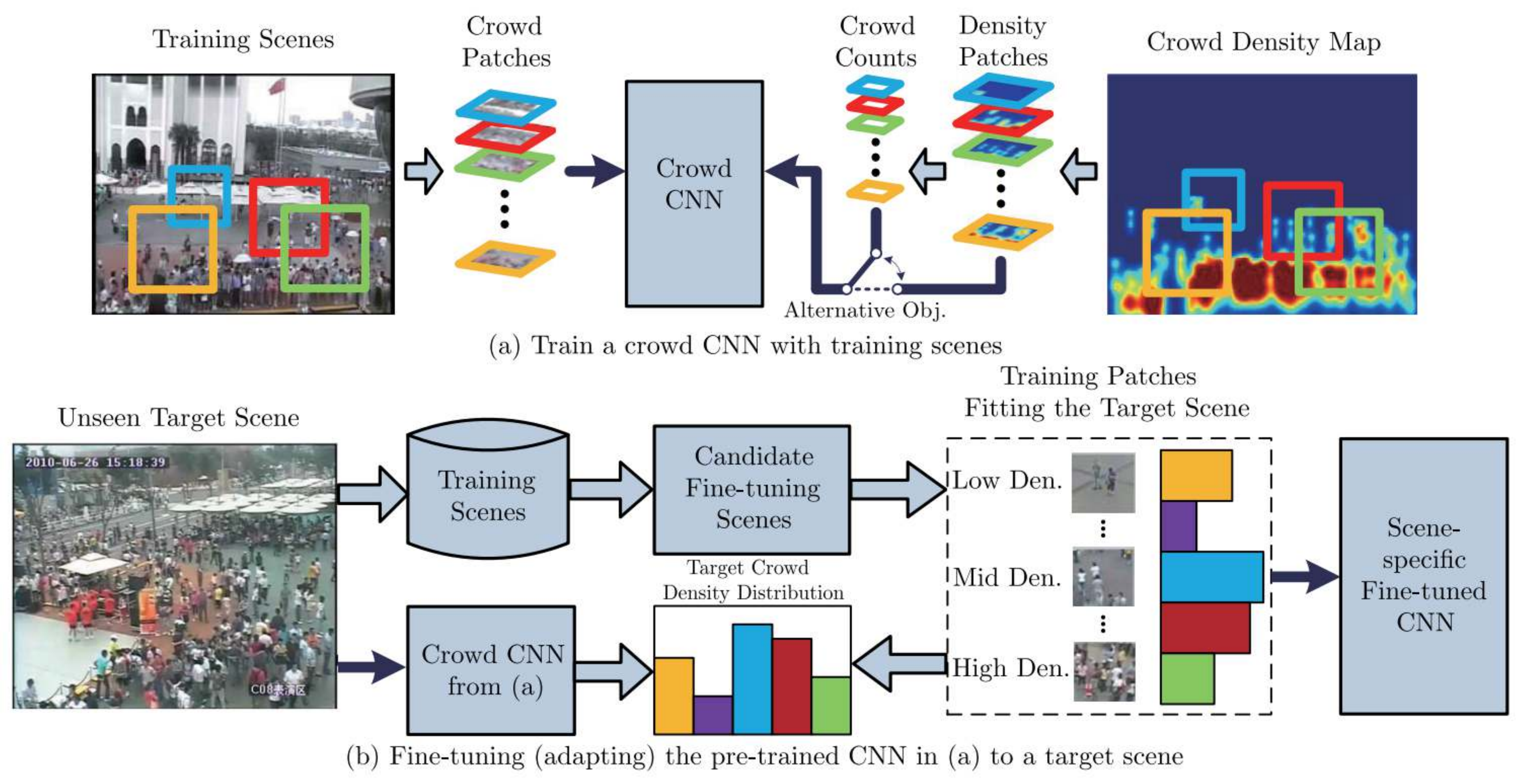}
\end{minipage}%
\vskip-8pt
\captionof{figure}{Overview of cross scene crowd counting proposed by Zhang \etal \cite{zhang2015cross}.}
\label{fig:cross_scene}
\end{center}
\end{figure}

Zhang \etal \cite{zhang2015cross} analyzed existing methods to identify that their performance reduces drastically when applied to a new scene that is different from the training dataset. To overcome this issue, they proposed to learn a mapping from images to crowd counts and to adapt this mapping to new target scenes for cross-scene counting. To achieve this, they first learned their network by alternatively training on two objective functions: crowd count and density estimation which are related objectives. By alternatively optimizing over these objective functions one is able to obtain better local optima. In order to adapt this network to a new scene, the network is  fine-tuned using training samples that are similar to the target scene. It is important to note that the network is adapted to new target scenes without any extra label information. The overview of their approach is shown in Fig. \ref{fig:cross_scene}. Also, in contrast to earlier methods that use the sum of Gaussian kernels centered on the locations of objects, a new method for generating ground truth density map is proposed that incorporates perspective information. In doing so, the network is able to perform perspective normalization thereby achieving robustness to scale and perspective variations. Additionally, they introduced a new dataset for the purpose of evaluating cross-scene crowd counting. The network is evaluated for cross-scene crowd counting as well as single scene crowd counting and superior results are demonstrated for both scenarios.

\begin{figure}[t]
\begin{center}
\begin{minipage}{0.7\linewidth}
\includegraphics[width=1\linewidth]{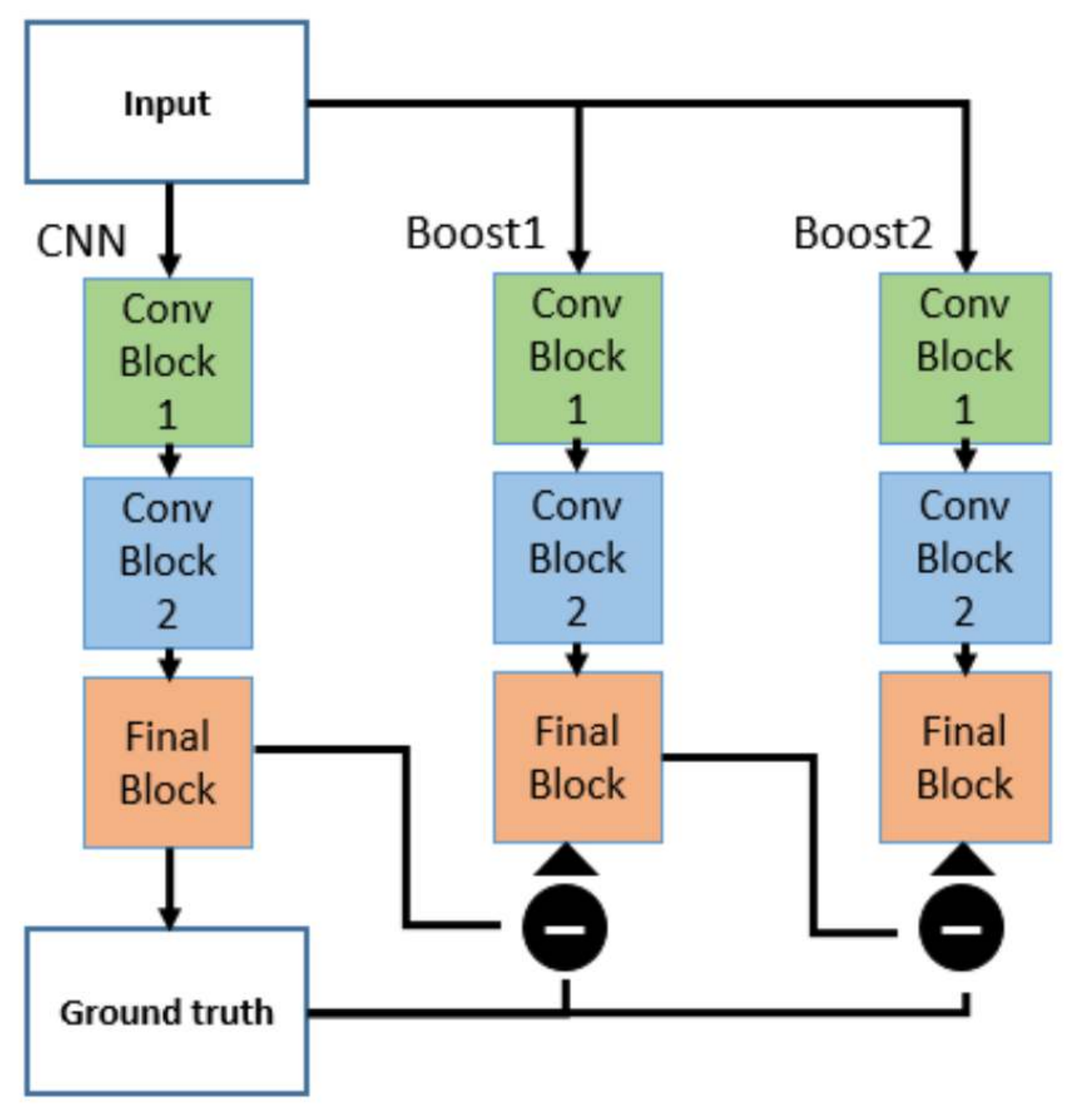}
\end{minipage}%
\vskip-8pt
\captionof{figure}{Overview of learning to count using boosting by Walach and Wolf \cite{walach2016learning}.}
\label{fig:learning_to_count_arch}
\end{center}
\end{figure}

Inspired by the success of cross-scene crowd counting \cite{zhang2015cross}, Walach and Wolf \cite{walach2016learning} performed layered boosting and selective sampling. Layered boosting involves iteratively adding CNN layers to the model such that every new layer is trained to estimate the residual error of the earlier prediction. For instance, after the first CNN layer is trained, the second CNN layer is trained on the difference between the estimation and ground truth. This layered boosting approach is based on the notion of Gradient Boosting Machines (GBM) \cite{friedman2001greedy} which are a subset of powerful ensemble techniques.  An overview of their boosting approach is presented in Fig. \ref{fig:learning_to_count_arch}. The other contribution made by the authors is the use of sample selection algorithm to improve the training process by reducing the effect of low quality samples such as trivial samples or outliers. According to the authors, the samples that are correctly classified early on are trivial samples. Presenting such samples for training even after the networks have learned to classify them tends to introduce bias in the network for such samples, thereby affecting its generalization performance. Another source of training inefficiency is the presence of outliers such as mislabeled samples. Apart from affecting the network's performance, these samples increase the training time. To overcome this issue, such samples are eliminated out of the training process for a number of epochs. The authors demonstrated that their method reduces the count estimation error by 20\% to 30\%  over existing state-of-the-art methods at that time on different datasets. 

\begin{figure}[htp!]
\begin{center}
\begin{minipage}{1\linewidth}
\includegraphics[width=\linewidth]{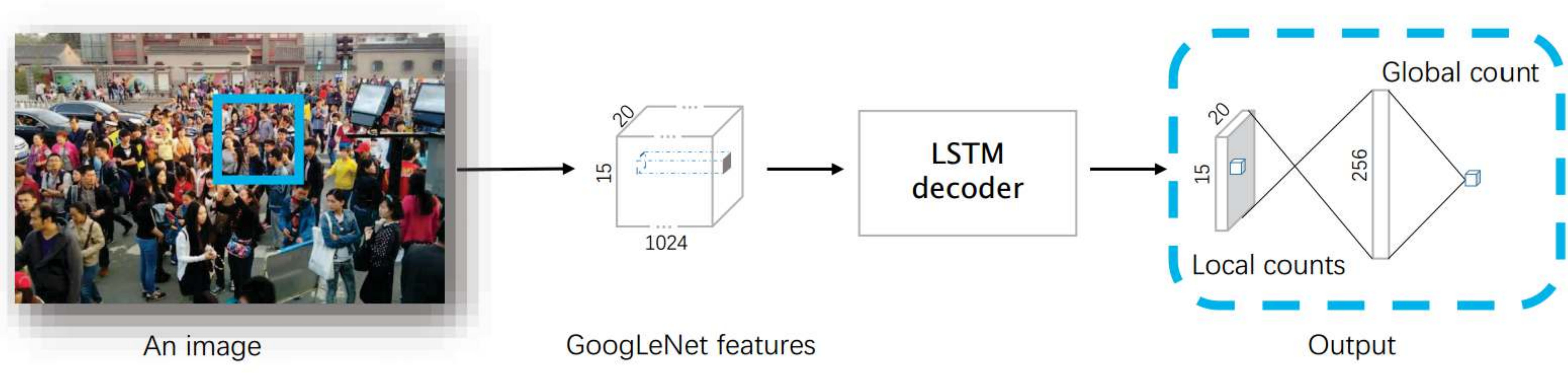}
\end{minipage}%
\vskip-8pt
\captionof{figure}{Overview of the end-to-end counting method proposed by Shang \etal \cite{skaug2016end}.  GoogLeNet is used to compute high-dimensional features which are further decoded into local counts using LSTM units. }
\label{fig:end_to_end_arch}
\end{center}
\end{figure}

In contrast to the above methods that use patch-based training, Shang \etal \cite{skaug2016end} proposed an end-to-end count estimation method using CNNs (Fig. \ref{fig:end_to_end_arch}). Instead of dividing the image into patches, their method takes the entire image as input and directly outputs the final crowd count. As a result, computations on overlapping regions are shared by combining multiple stages of processing leading to a reduction of complexity. The network simultaneously learns to estimate local counts and can be viewed as learning a patch level counting model which enables faster training. By doing so, contextual information is incorporated into the network, enabling it to ignore background noises and achieve better performance. The network is composed of three parts: (1) Pre-trained GoogLeNet model \cite{szegedy2015going}, (2) Long-short time memory (LSTM) decoders for local count, and (3) Fully connected layers for the final count. The network takes an image as input and computes high-dimensional  CNN feature maps using the GoogleNet network. Local blocks in these high-dimensional features are decoded into local count using a LSTM unit. A set of fully connected layers after the LSTM unit map the local counts into global count. The two counting objectives are jointly optimized during training.

\begin{figure}[htp!]
\begin{center}
\begin{minipage}{1\linewidth}
\includegraphics[width=\linewidth]{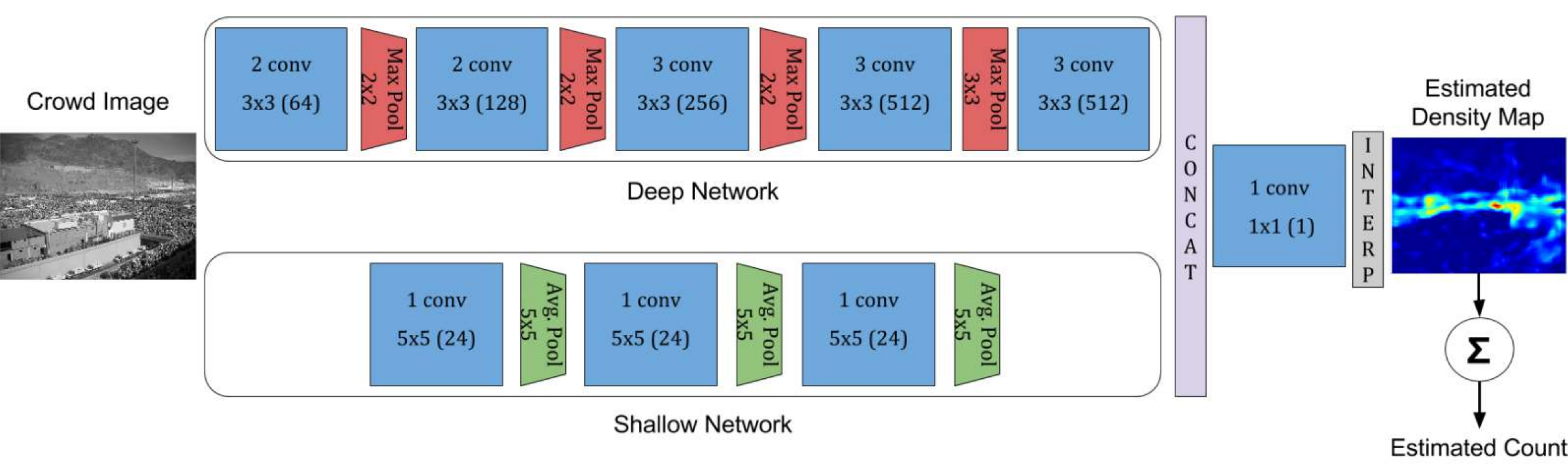}
\end{minipage}%
\vskip-8pt
\captionof{figure}{Overview of counting method proposed by Boominathan \etal \cite{boominathan2016crowdnet}.  A deep network is used in combination with a shallow network to address scale variations across images. }
\label{fig:boominathan_arch}
\end{center}
\end{figure}

In an effort to capture semantic information in the image, Boominathan \etal \cite{boominathan2016crowdnet} combined deep and shallow  fully convolutional networks to predict the density map for a given crowd image. The combination of two networks enables one to build a model robust to non-uniform scaling of crowd and variations in perspective. Furthermore, an extensive augmentation of the training dataset is performed in two ways. Patches from the multi-scale image representation are sampled to make the system robust to scale variations. Fig. \ref{fig:boominathan_arch} shows overview of this method. 

\begin{figure}[htp!]
\begin{center}
\begin{minipage}{1\linewidth}
\includegraphics[width=\linewidth]{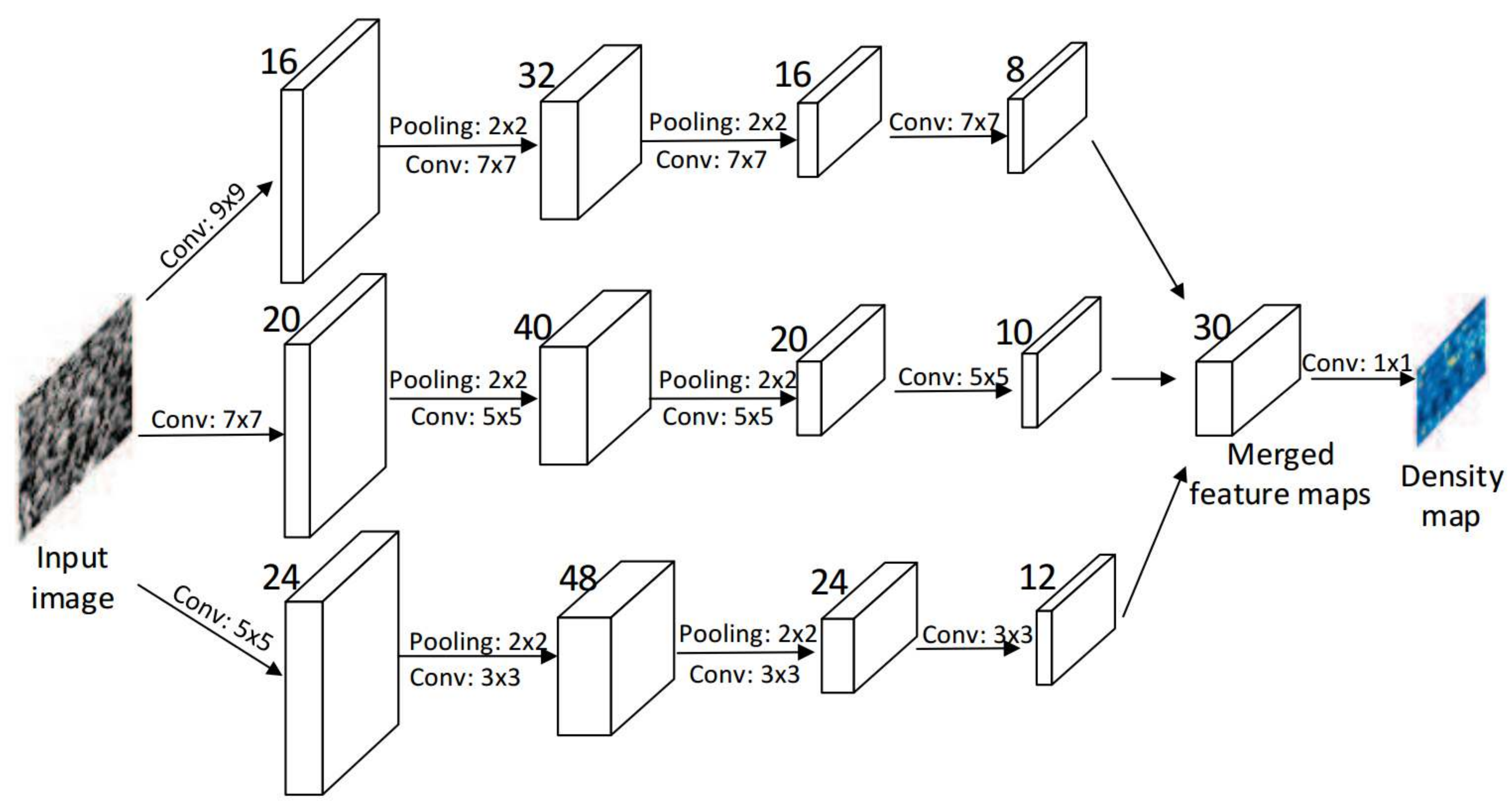}
\end{minipage}%
\vskip-8pt
\captionof{figure}{Overview of single image crowd counting via multi-column network  by Zhang \etal \cite{zhang2016single}.}
\label{fig:single_image_arch}
\end{center}
\end{figure}

In another approach, Zhang \etal \cite{zhang2016single} proposed a  multi-column based architecture (MCNN) for images with arbitrary crowd density and arbitrary perspective. Inspired by the success of multi-column networks for image recognition \cite{ciregan2012multi}, the proposed method ensures robustness to large variation in object scales by constructing a network that comprises of three columns corresponding to filters with receptive fields of different sizes (large, medium, small) as shown in Fig. \ref{fig:single_image_arch}. These different columns are designed to cater to different object scales present in the images. Additionally, a new method for generating ground truth crowd density maps is proposed. In contrast to existing methods that either use sum of Gaussian kernels with a fixed variance or perspective maps, Zhang \etal proposed to take into account perspective distortion by estimating spread parameter of the Gaussian kernel based on the size of the head of each person within the image. However, it is impractical to estimate head sizes and their underlying relationship with density maps. Instead they used an important property observed in high density crowd images that the head size is related to distance between the centers of two neighboring persons. The spread parameter for each person is data-adaptively determined based on its average distance to its neighbors. Note that the ground truth density maps created using this technique incorporate distortion information without the use of perspective maps. Finally, considering that existing crowd counting datasets do not cater to all the challenging situations encountered in real world scenarios, a new ShanghaiTech crowd datasets is constructed. This new dataset includes 1198 images with about 330,000 annotated heads.

\begin{figure}[htp!]
\begin{center}
\begin{minipage}{1\linewidth}
\includegraphics[width=\linewidth]{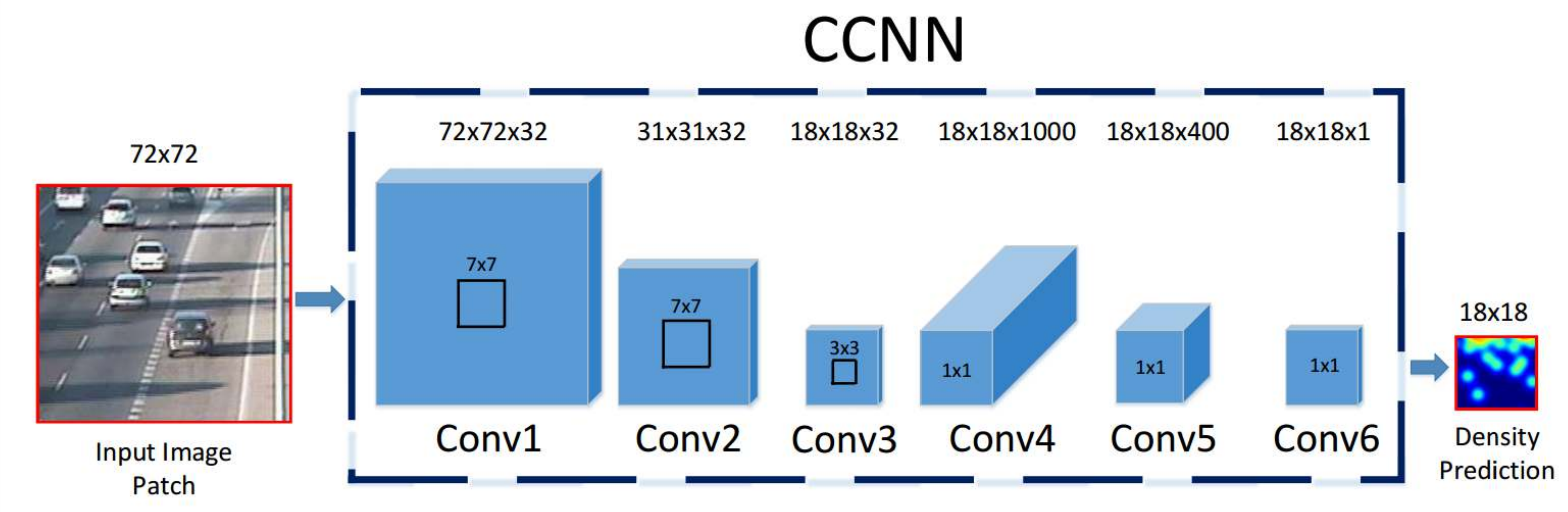}
\includegraphics[width=\linewidth]{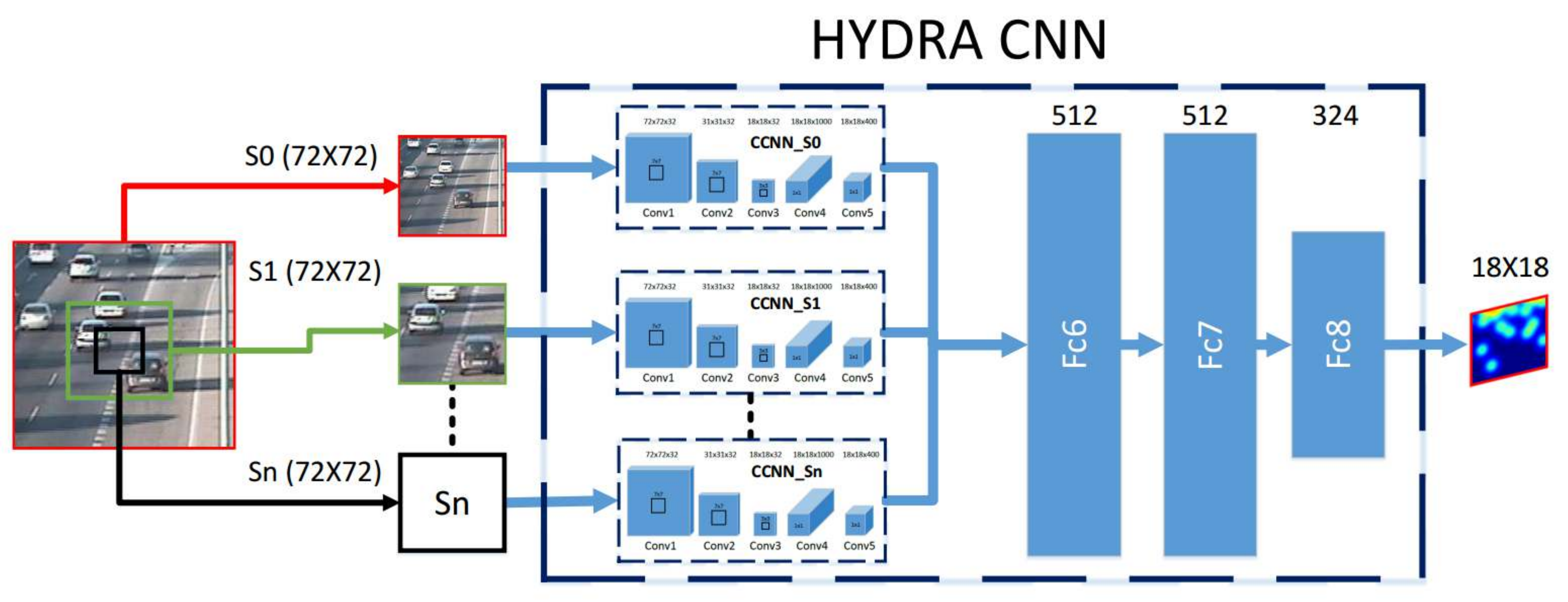}
\end{minipage}%
\vskip-10pt
\captionof{figure}{Overview of Hydra-CNN by Onoro \etal  \citep{onoro2016towards}.}
\label{fig:hydra_cnn_arch}
\end{center}
\end{figure}

Similar to the above approach, Onoro and Sastre \citep{onoro2016towards} developed a scale aware counting model called Hydra CNN that is able to estimate object densities in a variety of crowded scenarios without any explicit geometric information of the scene. First, a deep fully-convolutional neural network (which they call as Counting CNN) with six  convolutional layers is employed.  Motivated by the observation of earlier work \cite{zhang2015cross,loy2013crowd} that incorporating perspective information for geometric correction of the input features results in better accuracy, geometric information is incorporated into the Counting CNN (CCNN). To this effect, they developed Hydra CNN that learns a multi-scale non-linear regression model. As shown in Fig. \ref{fig:hydra_cnn_arch} the network consists of 3 heads and a body with each head learning features for a particular scale.  Each head of the Hydra-CNN is constructed using the CCNN model whose outputs are concatenated and fed to the body. The body consists of a set of two fully-connected layers followed by a rectified linear unit (ReLu), a dropout layer and a final fully connected layer to estimate the object density map. While the different heads extract image descriptors at different scales, the body learns a high-dimensional representation that fuses the multi-scale information provided by the heads. This network design of Hydra CNN is inspired by the work of Li et al. \cite{li2015visual}. Finally, the network is trained with  pyramid of image patches extracted at multiple scales. The authors demonstrated through their experiments that the Hydra CNN is able to perform successfully in scenarios and datasets with significant variations in the scene.

\begin{figure}[htp!]
\begin{center}
\begin{minipage}{1\linewidth}
\includegraphics[width=\linewidth]{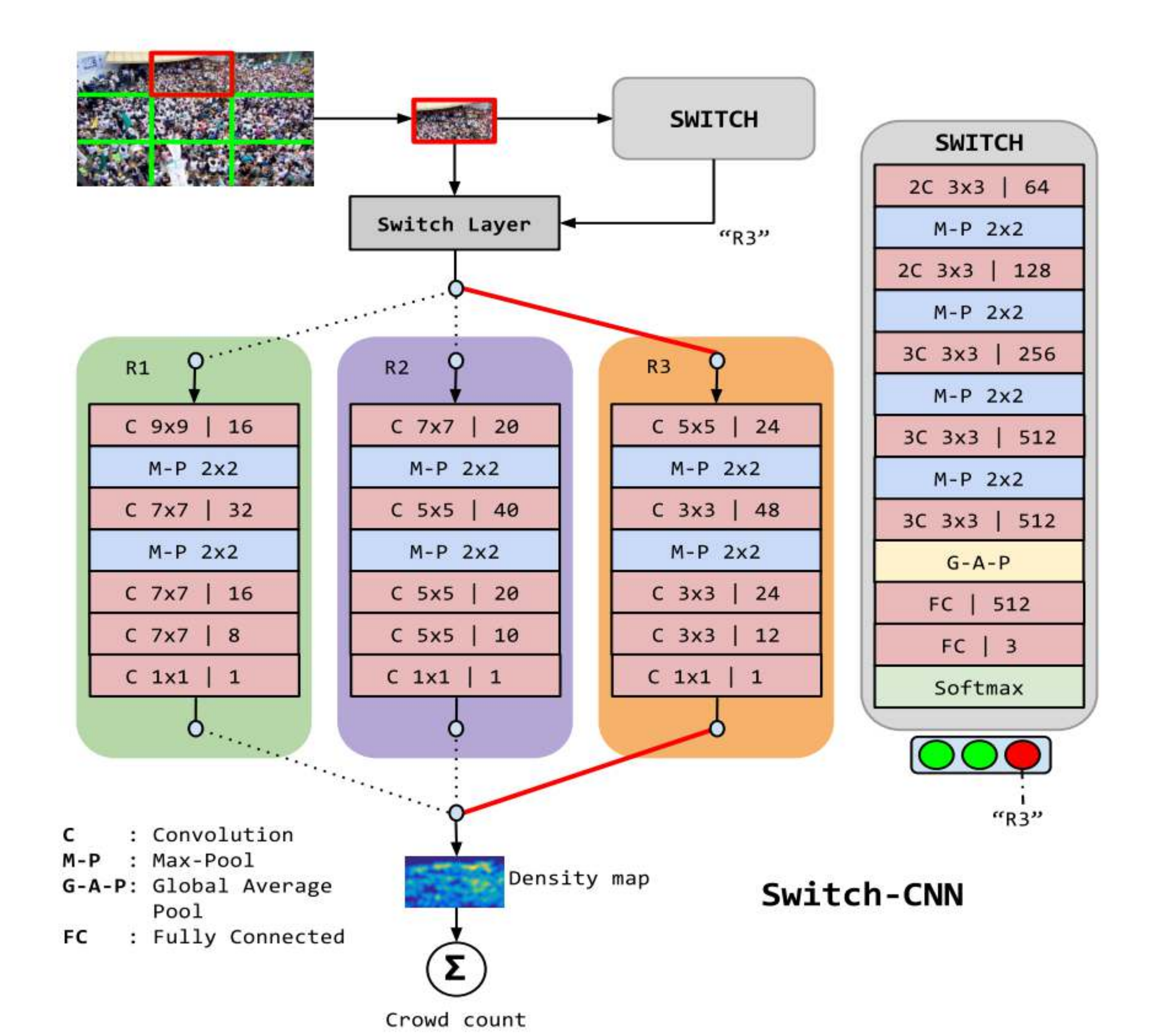}
\end{minipage}%
\vskip-8pt
\captionof{figure}{Overview of Switching CNN by Sam \etal \cite{sam2017switching}.}
\label{fig:switching_cnn}
\end{center}
\end{figure}

Instead of training all regressors of a multi-column network \cite{zhang2016single} on all the input patches, Sam \etal \cite{sam2017switching} argue that better performance is obtained by training regressors with a particular set of training patches by leveraging variation of crowd density within an image. To this end, they proposed a switching CNN that cleverly selects an optimal regressor suited for a particular input patch. As shown in Fig. \ref{fig:switching_cnn}, the proposed network consists of multiple independent regressors similar to multi-column network     \cite{zhang2016single} with different receptive fields and a switch classifier. The switch classifier is trained to select the optimal regressor for a particular input patch. Independent CNN crowd density regressors are trained on patches sampled from a grid in a given crowd scene. The switch classifier and the independent regressors are alternatively trained. The authors describe multiple stages of training their network. First, the independent regressors are pretrained on image patches to minimize the Euclidean distance between the estimated density map and ground truth. This is followed by a differential training stage where, the count error is factored in to improve the counting performance by back-propagating a regressor with the minimum count error for a given training patch. After training the multiple regressors, a switch classifier based on VGG-16 architecture \cite{simonyan2014very} is trained to select an optimal regressor for accurate counting. Finally, the switch classifier and CNN regressors are co-adapted in the coupled training stage.

While the above methods concentrated on incorporating scale information in the network, Sheng \etal in \cite{sheng2016crowd} proposed to integrate semantic information by learning locality-aware feature sets. Noting that earlier methods that use hand-crafted features ignored key semantic  and spatial information, the authors proposed a new image representation which incorporates semantic attributes as well as spatial cues to improve the discriminative power of feature representations. They defined semantic attributes at the pixel level and learned  semantic feature maps via deep CNN. The spatial information in the image is encoded using locality-aware features in the semantic attribute feature map space. The locality-aware features (LAF) are built on the idea of spatial pyramids on neighboring patches thereby encoding spatial context and local information.  The local descriptors from adjacent cells are then encoded into image representations using weighted VLAD encoding method.

\begin{figure}[htp!]
\begin{center}
\begin{minipage}{.8\linewidth}
\includegraphics[width=\linewidth]{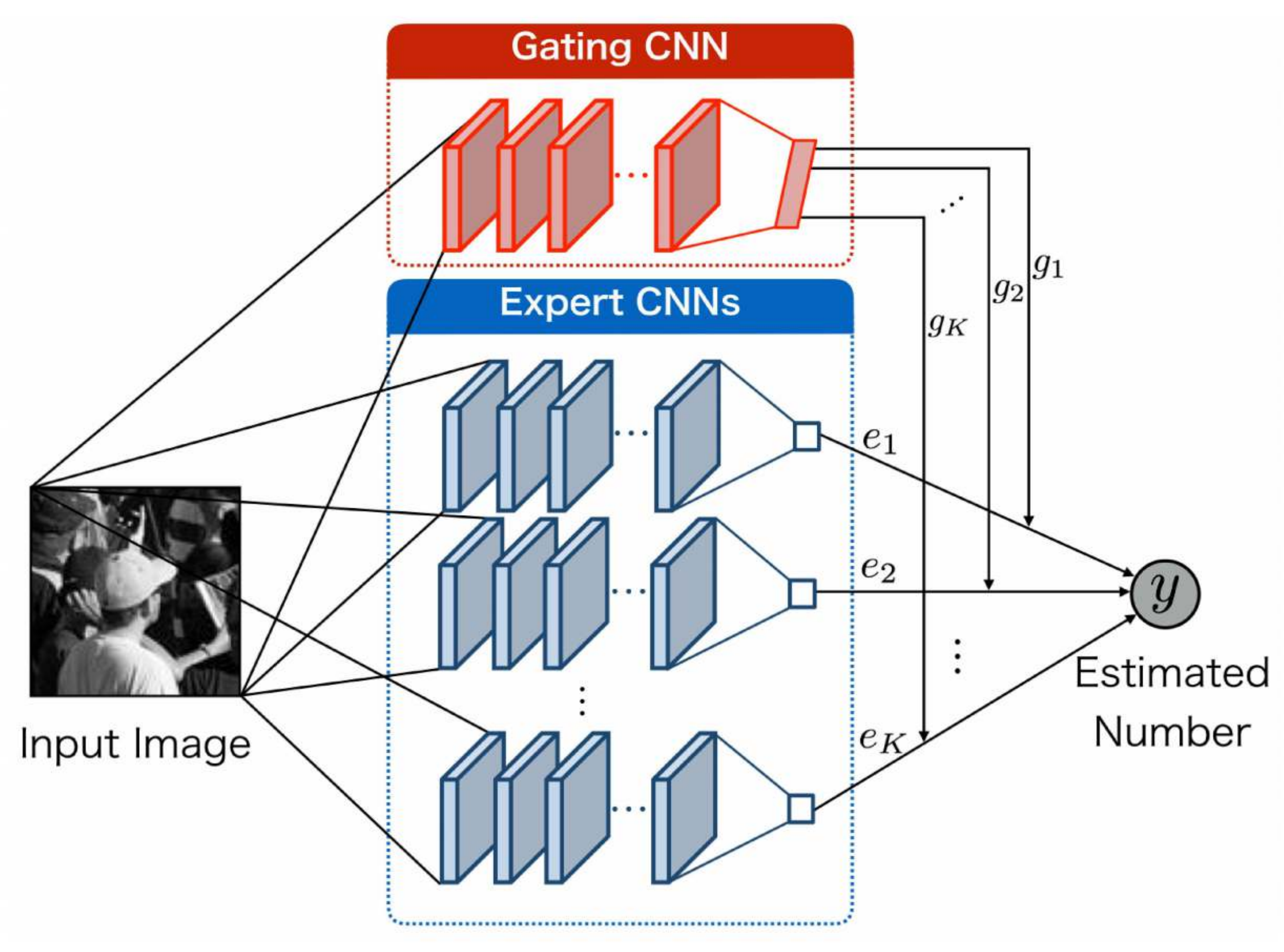}
\end{minipage}%
\vskip-8pt
\captionof{figure}{Overview of MoC (Mixture of CNN) for crowd counting by Kumagai \etal \cite{kumagai2017mixture}.}
\label{fig:mixture_cnn_arch}
\end{center}
\end{figure}

Similar to \cite{zhang2016single,onoro2016towards}, Kumagai \etal \cite{kumagai2017mixture}, based on the observation that a single predictor is insufficient to appropriately predict the count in the presence of large appearance changes, proposed a Mixture of CNNs (MoCNN) that are specialized to a different scene appearances. As shown in Fig. \ref{fig:mixture_cnn_arch}, the architecture consists of a mixture of expert CNNs and a gating CNN that adaptively selects the appropriate CNN among the experts according to the appearance of the input image. For prediction, the expert CNNs predict crowd count in the image while the gating CNN predicts appropriate probabilities for each of the expert CNNs. These probabilities are further used as weighting factors to compute the weighted average of the counts predicted by all the expert CNNs.

Motivated by the success of scale aware models \cite{zhang2016single,onoro2016towards}, Marsden \etal \cite{marsden2016fully} proposed to incorporate scale into the models with much less number of model parameters.  Observing that the earlier scale aware models \cite{zhang2016single,onoro2016towards} are difficult to optimize and are computationally complex, Marsden \etal \cite{marsden2016fully} proposed a single column fully convolutional network where the scale information is incorporated into the model using a simple yet effective multi-scale averaging step during prediction without any increase in the model parameters. The method addresses the issues of scale and perspective changes by feeding multiple scales of test image into the network during prediction phase. The crowd count is estimated for each scale and the final count is obtained by taking an average of all the estimates. Additionally,  a new training set augmentation scheme is developed to reduce redundancy among the training samples. In contrast to the earlier methods that use  randomly cropped patches with high degree of overlap, the training set in this work is constructed using the four image quadrants as well as their horizontal flips ensuring no overlap. This technique avoids potential overfit when the network is continuously exposed to the same set of pixels during training, thereby improving the generalization performance of the network. In addition, the generalization performance of the proposed method is studied by measuring cross dataset performance.

\begin{figure}[htp!]
\begin{center}
\begin{minipage}{0.8\linewidth}
\includegraphics[width=\linewidth]{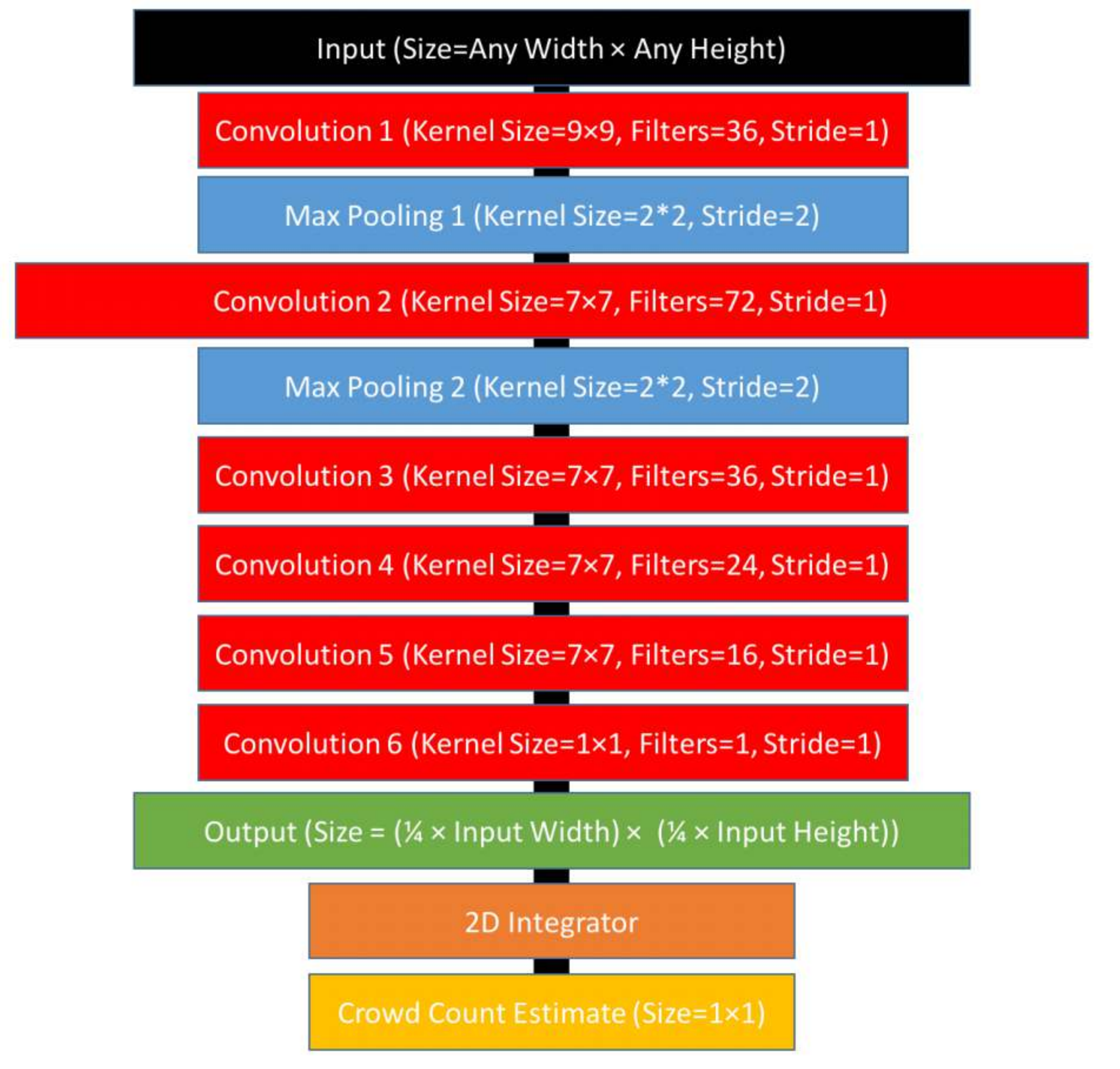}
\end{minipage}%
\vskip-8pt
\captionof{figure}{Overview of Fully Convolutional Network  for crowd counting by Marsden \etal \cite{marsden2016fully}.}
\label{fig:congested_arch}
\end{center}
\end{figure}

\begin{figure}[htp!]
\begin{center}
\begin{minipage}{1\linewidth}
\includegraphics[width=\linewidth]{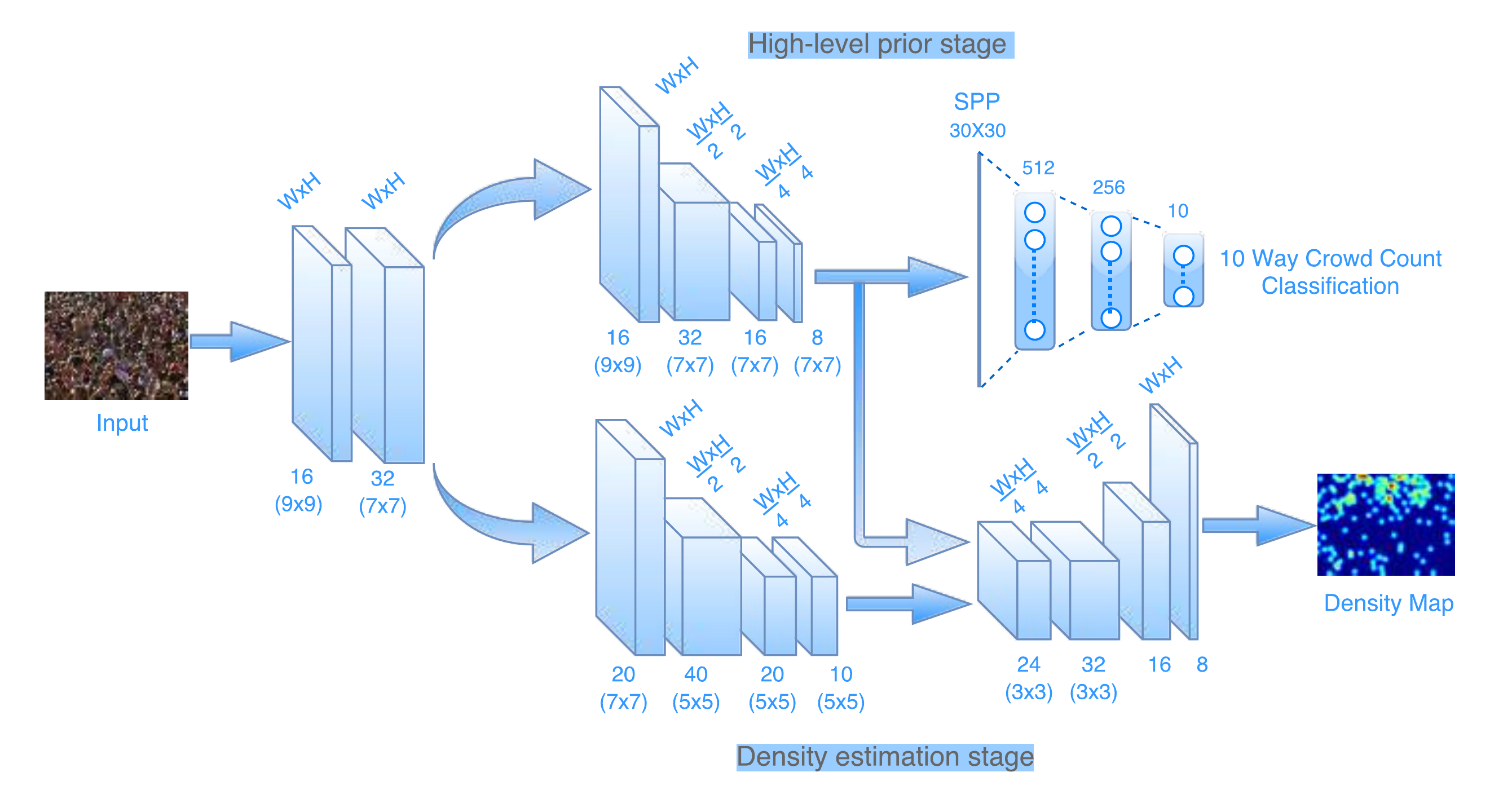}
\end{minipage}%
\vskip-8pt
\captionof{figure}{Overview of Cascaded Multi-task CNN by Sindagi \etal \cite{sindagi2017cnnbased}.}
\label{fig:cascaded_mtcnn}
\end{center}
\end{figure}

Inspired by the superior results achieved by simultaneous learning of related tasks \cite{ranjan2016hyperface,yu2017iprivacy}, Sindagi \etal \cite{sindagi2017cnnbased} and Marsden et al.  \cite{marsden2017resnetcrowd} explored multi-task learning to boost individual task performance. Marsden et al. \cite{marsden2017resnetcrowd} proposed a Resnet-18 \cite{he2016deep} based architecture for simultaneous crowd counting, violent behaviour detection and crowd density level classification. The network consists of initial 5 convolutional layers of Resnet18 including batch normalisation layers and skip connections form the primary module. The convolutional layers are followed by a set of task specific layers. Finally, sum of all the losses corresponding to different tasks is minimized. Additionally, the authors constructed a new 100 image dataset specifically designed for multi-task learning of crowd count and behaviour. In a different approach, Sindagi \etal \cite{sindagi2017cnnbased} proposed a cascaded CNN architecture to incorporate learning of a high-level prior to boost the density estimation performance. Inspired by \cite{chen2016cascaded}, the proposed network simultaneously learns to classify the crowd count into various density levels and estimate density map (as shown in Fig. \ref{fig:cascaded_mtcnn}). Classifying crowd count into various levels is equivalent to coarsely estimating the total count in the image thereby incorporating a high-level prior into the density estimation network. This enables the layers in the network to learn globally relevant discriminative features. Additionally, in contrast to most recent work, they make use of transposed convolutional layers to generate high resolution density maps.

In a recent work, Kang \etal \cite{kang2017beyond} explored maps generated by density estimation methods for the purpose of various crowd analysis tasks such as counting, detection and tracking. They performed a detailed analysis of the effect of using full-resolution density maps on the performance of these tasks. They demonstrated through their experiments that full resolution density maps improved the performance of localization  tasks such as detection and tracking. Two different approaches are considered for generating full-resolution maps. In the first approach, a sliding window based CNN regressor is used for pixel-wise density prediction. In the second approach, Fully Convolutional Networks \cite{long2015fully} along with skip connections are used to learning a non-linear mapping between input image and the corresponding density map.

In a slightly different application context of counting, Mundhenk \etal \cite{mundhenk2016large} and Arteta \etal \cite{arteta2016counting} proposed to count different types of objects such as cars and penguins respectively. Mundhenk \etal \cite{mundhenk2016large} addressed the problem of automated counting of automobiles from satellite/aerial platforms. Their primary contribution is the creation of a large diverse set of cars from overhead images. Along with the large dataset, they present a deep CNN-based network to recognize the number of cars in patches. The network is trained in a classification setting where the output of the network is a class that is indicative of the number of objects in the input image. Also, they incorporated contextual information by including additional regions around the cars in the training patches. Three different networks based on AlexNet \cite{krizhevsky2012imagenet}, GoogLeNet \cite{szegedy2015going} and ResNet \cite{he2016deep} with Inception are evaluated. For a different application of counting penguins in images, Arteta \etal \cite{arteta2016counting} proposed a deep multi-task architecture for accurate counting even in the presence of labeling errors. The network is trained in a multi-task setting where, the tasks of foreground-background subtraction and uncertainty estimation along with counting are jointly learned. The authors demonstrated that the joint learning especially helps in learning a counting model that is robust to labeling errors. Additionally, they exploited scale variations and count variability across the annotations  to incorporate scale information of the object and prediction of annotation difficulty respectively into the model. The network was evaluated on a newly created Penguin dataset.

\begin{figure}[htp!]
\begin{center}
\begin{minipage}{1\linewidth}
\includegraphics[width=\linewidth]{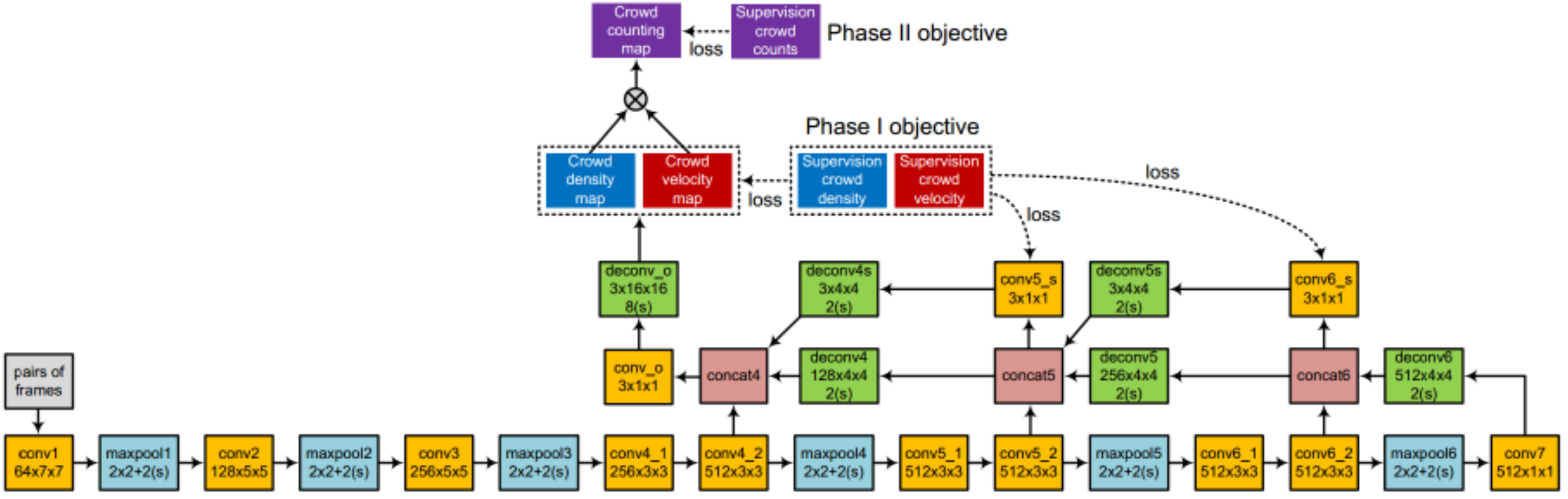}
\end{minipage}%
\vskip-8pt
\captionof{figure}{Overview of the method proposed by Zhao \etal \cite{zhao2016crossing} for counting people crossing a line.}
\label{fig:crossing_line_arch}
\end{center}
\end{figure}

Zhao \etal addressed a higher level cognitive task of counting people that cross a line in \citep{zhao2016crossing}. Though the task is a video-based application, it comprises of a CNN-based model that is trained with pixel-level supervision maps similar to single image crowd density estimation methods, making it a relevant approach to be included in this article. Their method consists of a two-phase training scheme (as shown in Fig. \ref{fig:crossing_line_arch}) that decomposes  original counting problem into two sub-problems: estimating crowd density map and crowd velocity map where  the two tasks share the initial set of layers enabling them to learn more effectively. The estimated crowd density and crowd velocity maps are then multiplied element-wise to generate the crowd counting maps. Additionally, they contributed a large-scale dataset for evaluating crossing-line crowd counting algorithms, which includes 5 different scenes, 3,100 annotated frames and 5,900 annotated pedestrians.

\section{Discussion}

With a variety of methods discussed in Section \ref{sec:survey_cnn}, we analyze various advantages and disadvantages of the broad approaches followed by these methods in this section. 

Zhang \etal \cite{zhang2015cross} were among the first ones to address the problem of adapting models to new unlabelled datasets using a simple and effective method based on finding similar patches across datasets. However, their method is heavily dependent on accurate perspective maps which may not be necessarily available for all the datasets. Additionally, the use of 72$\times$72 sized patches for training and evaluation ignores global context which is necessary for accurate estimation of count. Walach \etal \cite{walach2016learning} successfully addressed training inefficiencies in earlier methods using a layered boosting approach and a simple sample selection method. However, similar to Zhang \etal \cite{zhang2015cross}, their method involves patch-based training and evaluation resulting in loss of global context information along with inefficiency during evaluation due to the use of a sliding window approach. Additionally, these methods tend to ignore scale variance among the dataset assuming that their models will implicitly learn the invariance.

In an effort to explicitly model scale invariance, several methods involving combination of networks were proposed (\cite{zhang2016single,onoro2016towards,sam2017switching,kumagai2017mixture,boominathan2016crowdnet}). 
While these methods demonstrated significant improvements in the performance using multiple column networks and a combination of deep and shallow networks, the invariance achieved is limited by the number of columns present in the network and receptive field sizes which are chosen based on the scales present in the dataset. Additionally, these methods do not explicitly model global context information which is crucial for a task such as crowd counting. In a different approach, Marsden \etal \cite{marsden2016fully} attempt to address the scale issue by performing a multi-scale averaging during the prediction phase. While being simple and effective, it results in an inefficient inference stage. Additionally, these methods do not explicitly encode global context present in an image which can be crucial for improving the count performance. To this end, few approaches model local and global context \cite{sheng2016crowd,skaug2016end} by considering key spatial and semantic information present in the image. 

In an entirely different approach, few methods \cite{marsden2017resnetcrowd,sindagi2017cnnbased} take advantage of multi-task learning and incorporate high-level priors into the network. For instance, Sindagi \etal \cite{sindagi2017cnnbased} simultaneously learn density estimation and a high-level prior in the form of crowd count classification. While they demonstrated high performance gain by learning an additional task of crowd density level classification, the number of density levels is dataset dependent and it needs to be carefully chosen based on the density levels present in the dataset.

\section{Datasets and results}
\label{sec:datasets_and_results}
A variety of datasets have been created over the last few years driving researchers to create models with better generalization abilities. While the earlier datasets usually contain low density crowd images, the most recent ones focus on high density crowd thus posing numerous challenges such as scale variations, clutter and severe occlusion. The creation of these large scale datasets has motivated recent approaches to develop methods that cater to such challenges. In this section, we review five key datasets \cite{chan2008privacy,chen2012feature,idrees2013multi,zhang2015cross,zhang2016single} followed by a discussion on the results of CNN-based approaches and recent traditional methods that were not included in the earlier surveys. 

\begin{table*}[t!]
\caption{Summary of various datasets.}
\begin{center}
\vskip-15pt\begin{tabular}{|l|c|c|c|c|c|c|}
\hline
Dataset & No. of images & Resolution & Min & Ave & Max & Total count \\
\hline
UCSD \cite{chan2008privacy} & 2000 & 158x238 & 11 & 25 & 46 & 49,885\\
\hline
Mall \cite{chen2012feature} & 2000 & 320x240 & 13 & - & 53 & 62,325\\
\hline
UCF\textunderscore CC\textunderscore 50 \cite{idrees2013multi} & 50 & Varied & 94 & 1279 & 4543 & 63,974\\
\hline
WorldExpo '10 \cite{zhang2016data,zhang2015cross} & 3980 & 576x720 & 1 & 50 & 253 & 199,923\\
\hline
ShanghaiTech Part A \cite{zhang2016single} & 482 & Varied & 33 & 501 & 3139 & 241,677\\
\hline
ShanghaiTech Part B \cite{zhang2016single}& 716 & 768x1024 & 9 & 123 & 578 & 88,488\\
\hline
\end{tabular}
\end{center}
\label{tab:datasetsummary}
\end{table*}

\begin{figure*}[ht!]
\begin{center}
\begin{minipage}{0.163\linewidth}
\includegraphics[width=\linewidth]{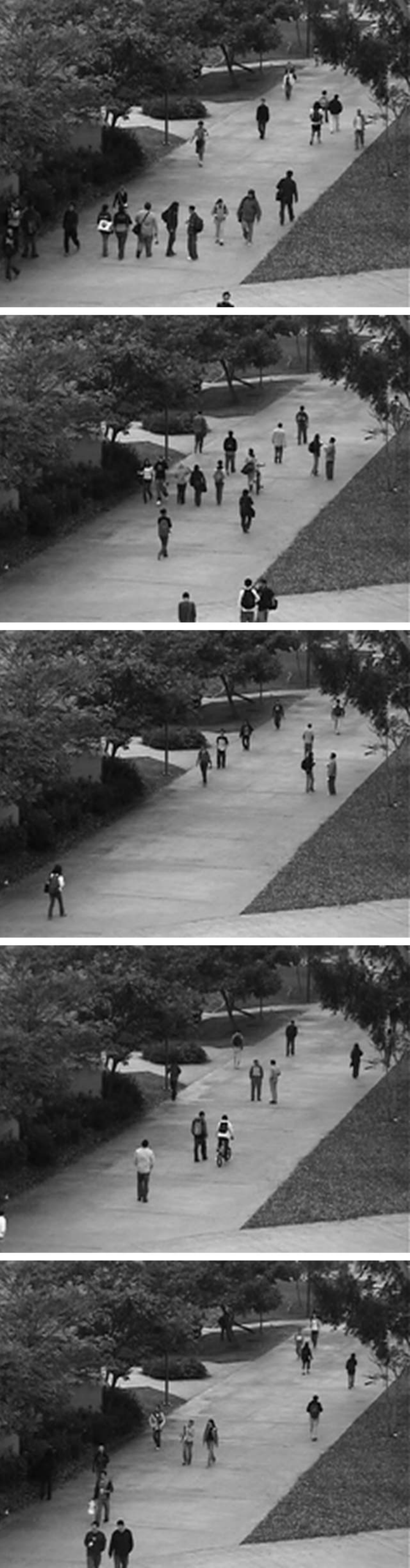}
\captionof*{figure}{(a)}
\end{minipage}%
\hfill
\begin{minipage}{0.163\linewidth}
\includegraphics[width=\linewidth]{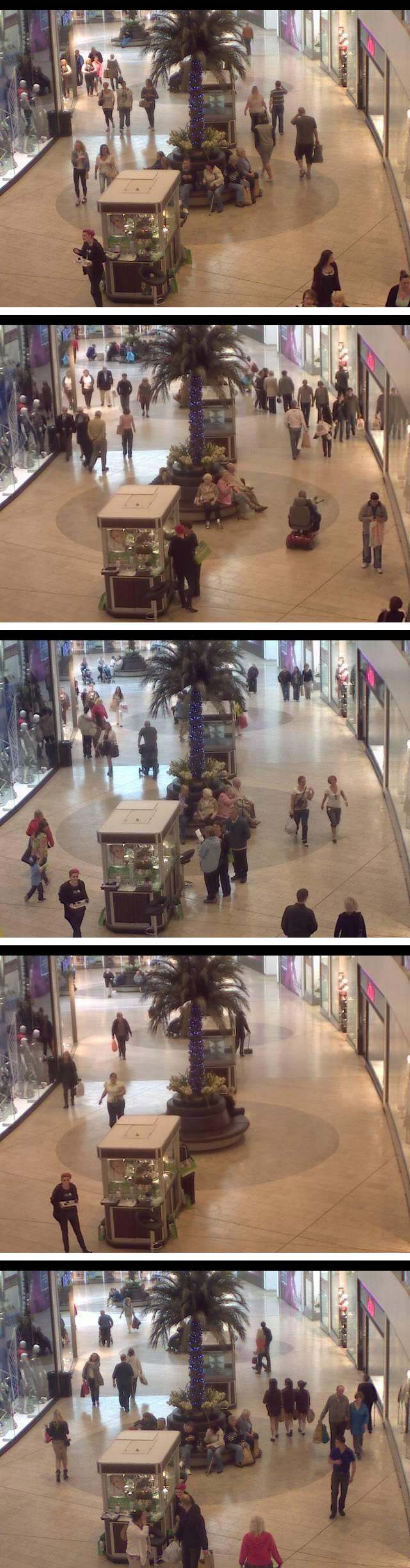}
\captionof*{figure}{(b)}
\end{minipage}
\begin{minipage}{0.163\linewidth}
\includegraphics[width=\linewidth]{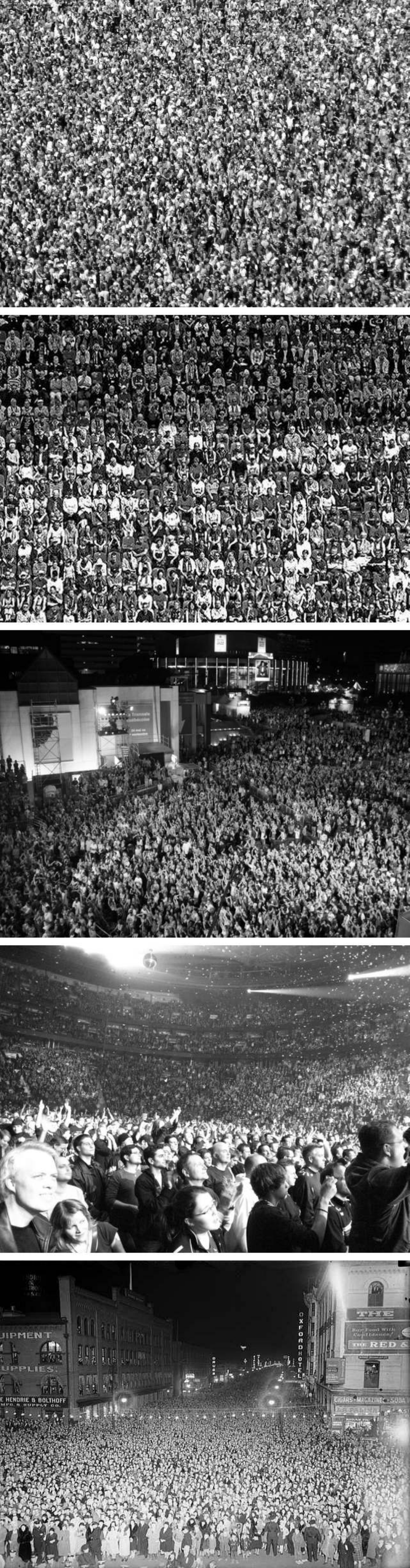}
 \captionof*{figure}{(c)}
\end{minipage}
\begin{minipage}{0.163\linewidth}
\includegraphics[width=\linewidth]{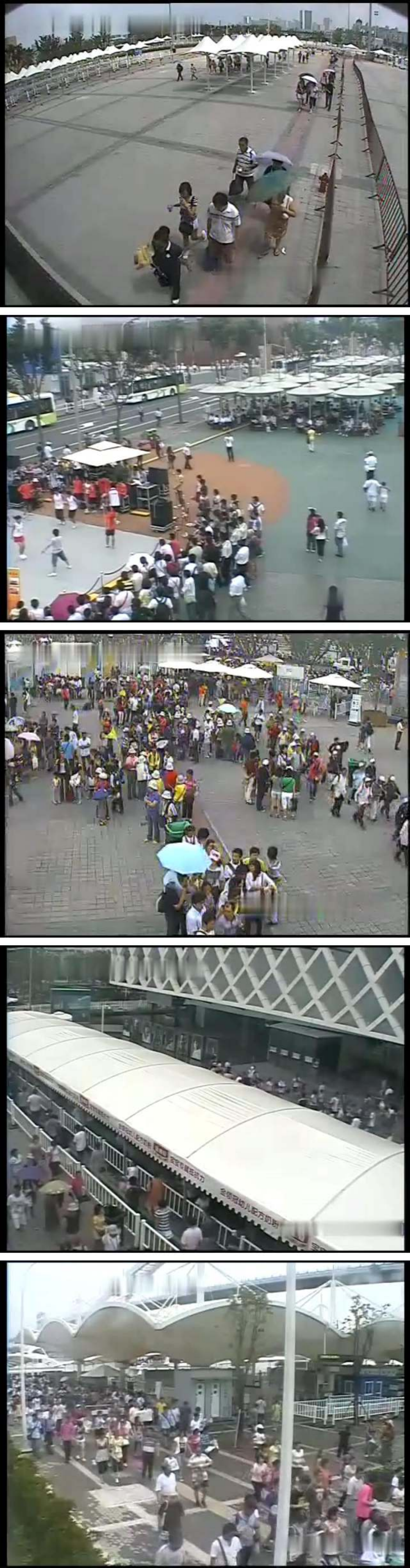}
\captionof*{figure}{(d)}
\end{minipage}
\begin{minipage}{0.163\linewidth}
\includegraphics[width=\linewidth]{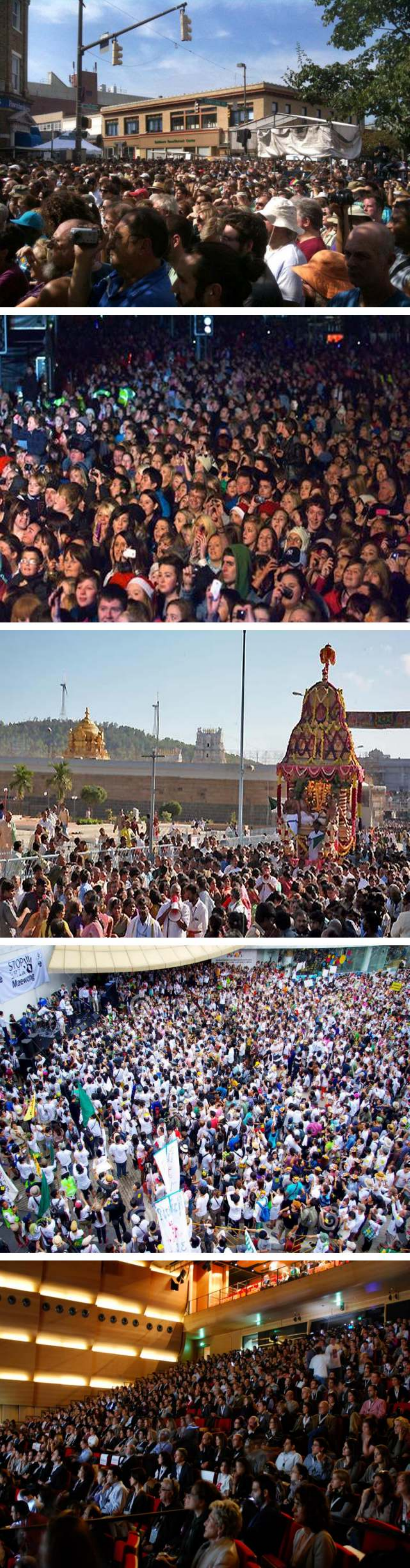}
\captionof*{figure}{(e)}
\end{minipage}
\begin{minipage}{0.163\linewidth}
\includegraphics[width=\linewidth]{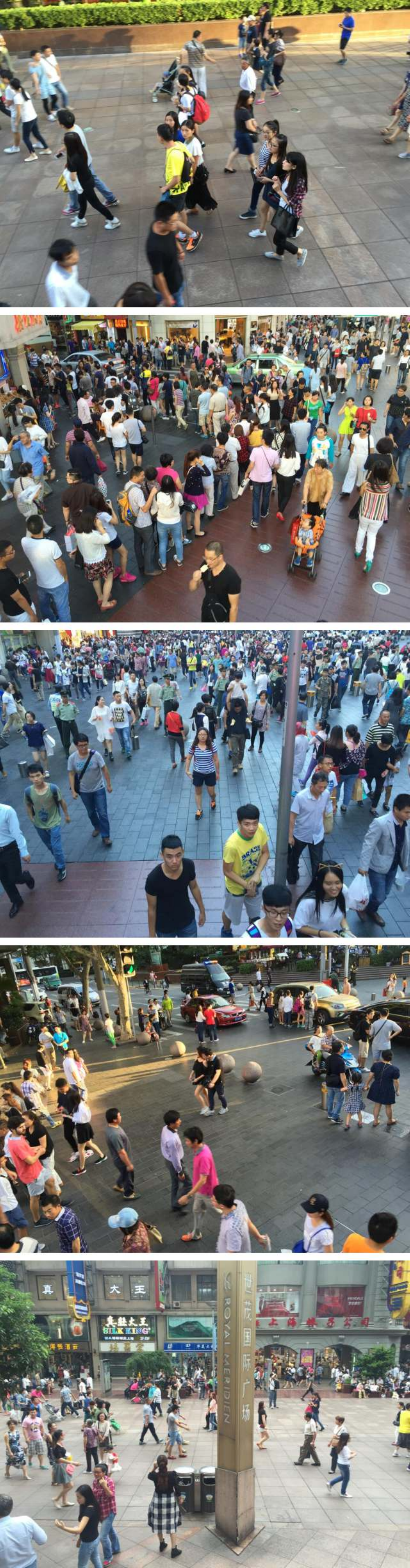}
\captionof*{figure}{(f)}
\end{minipage}
\end{center}
\vskip-10pt \captionof{figure}{Sample images from various datasets. (a) UCSD \cite{chan2008privacy} (b) Mall \cite{chen2012feature} (c)  UCF\textunderscore CC\textunderscore 50 \cite{idrees2013multi} (d) WorldExpo '10 \cite{zhang2015cross} (e) Shanghai Tech Part A \cite{zhang2016single} (f) SHanghai Tech Part B \cite{zhang2016single}. It can be observed that in the case of UCSD and Mall dataset , the images come from the same video sequence providing no variation in perspective across images.}
\label{fig:dataset}
\end{figure*}

\subsection{Datasets}
\textbf{UCSD dataset}: The UCSD dataset \cite{chan2008privacy} was among the first datasets to be created for counting people. The dataset was collected from a video camera at a pedestrian walkway. The dataset consists of 2000 frames of size 238$\times$158 from a video sequence along with ground truth annotations of each pedestrian in every fifth frame. For the rest of the frames, linear interpolation is used to create the annotations. A region-of-interest is also provided to ignore unnecessary moving objects such as trees. The dataset contains a total of 49,885 pedestrian instances and it is split into training and test set. While the training set contains frames with indices 600 to 1399, the test set contains the remaining 1200 images. This dataset has relatively low density crowd with an average of around 15 people in a frame and since the dataset was collected from a single location, there is no variation in the scene perspective across images. 
\linebreak

\noindent \textbf{Mall dataset}: Considering little variation in the scene type in the UCSD dataset, Chen \etal in \cite{chen2012feature} collected a new Mall dataset with diverse illumination conditions and crowd densities. The dataset was collected using a surveillance camera installed in a shopping mall. Along with having various density levels, it also has different activity patterns (static and moving crowds). Additionally, the scene contained in the dataset has severe perspective distortion resulting in large variations in size and appearance of objects. The dataset also presents the challenge of severe occlusions caused by the scene objects, e.g.stall, indoor plants along the walking path. The video sequence in the dataset consists of 2000 frames of size 320$\times$240 with 6000 instances of labelled pedestrians. The first 800 frames are used for training and the remaining 1200 frames are used for evaluation. In comparison to the UCSD dataset, the Mall dataset has relatively higher crowd density images. However, both the datasets do not have any variation in the scene perspective across images since they are a part of a single continuous video sequence. 
\linebreak

\noindent \textbf{UCF\textunderscore CC\textunderscore 50 dataset}: The UCF\textunderscore CC\textunderscore 50 \cite{idrees2013multi} is the first truly challenging dataset constructed to include a wide range of densities and diverse scenes with varying perspective distortion. The dataset was created from publicly available web images. In order to capture diversity in the scene types, the authors collected images with different tags such as concerts, protests, stadiums and marathons. It contains a total of 50 images of varying resolutions with an average of 1280 individuals per image. A total of 63075 individuals were labelled in the entire dataset. The number of individuals varies from 94 to 4543 indicating a large variation across the images. The only drawback of this dataset is that only a limited number of images are available for training and evaluation. Considering the low number of images, the authors defined a cross-validation protocol for training and testing their approach where the dataset was divided into sets of 10 and a five fold cross-validation is performed. The challenges posed by this dataset are so enormous that even the results of recent CNN-based state-of-the-art approaches on this dataset are far from optimal.
\linebreak

\noindent \textbf{WorldExpo '10 dataset}: Since some of the earlier approaches and datasets focussed primarily on single scene counting, Zhang \etal \cite{zhang2015cross} introduced a dataset for the purpose of cross-scene crowd counting.  The authors attempted to perform a data-driven cross-scene crowd counting for which they collected a new large-scale dataset that includes 1132 annotated video sequences captured by 108 surveillance cameras, all from Shanghai 2010 WorldExpo event. Large diversity in the scene types is ensured by collecting videos from cameras having disjoint bird views. The dataset consists of a total of 3980 frames of size 576 $\times$ 720 with 199923 labelled pedestrians. The dataset is split into two parts: training set consisting of 1,127 one-minute long video sequences from 103 scenes and test set consisting of 5 one-hour long video sequences from 5 different scenes. Each test scene consists of 120 labelled frames with the crowd count varying from 1 to 220. Though an attempt is made to capture diverse scenes with varying density levels, the diversity is limited to only 5 scenes in the test set and the maximum crowd count is limited to 220. Hence, the dataset is not sufficient enough for evaluating  approaches designed for extremely dense crowds in a variety of scenes. 
\linebreak

\begin{table*}[htp!]
\centering
\caption{Comparison of results on various datasets. The CNN-based approaches provide significant improvements over traditional approaches that rely on hand-crafted representations. Further, among the CNN-based methods, scale aware and context aware approaches tend to achieve lower count error.}
\label{tab:results}
\resizebox{0.70\textwidth}{!}{%
\begin{tabular}{|c|l|c|c|c|c|c|c|c|c|c|c|c|c|}
\hline
\multicolumn{1}{|l|}{}                                                                        & Dataset                                                                                                   & \multicolumn{2}{c|}{UCSD}                 & \multicolumn{2}{c|}{Mall}                 & \multicolumn{2}{c|}{UCF CC 50}              & \multicolumn{2}{c|}{\begin{tabular}[c]{@{}c@{}}WorldExpo\\  '10\end{tabular}} & \multicolumn{2}{c|}{\begin{tabular}[c]{@{}c@{}}Shanghai\\ Tech-A\end{tabular}} & \multicolumn{2}{c|}{\begin{tabular}[c]{@{}c@{}}Shanghai\\ Tech-B\end{tabular}} \\ \hline
\multicolumn{1}{|l|}{\begin{tabular}[c]{@{}l@{}}Approach\\ type\end{tabular}}                 & Method                                                                                                    & MAE                 & MSE                 & MAE                 & MSE                 & MAE                  & MSE                  & MAE                                           & MSE                           & MAE                                    & MSE                                   & MAE                                    & MSE                                   \\ \hline
\multirow{5}{*}{\rotatebox[origin=c]{90}{\parbox[c]{4.5cm}{\centering Traditional approaches}}} & \begin{tabular}[c]{@{}l@{}}Multi-source multi-scale\\ Idrees \etal \cite{idrees2013multi}\end{tabular}    &                     &                     &                     &                     & 468.0                & 590.3                &                                               &                               &                                        &                                       &                                        &                                       \\ \cline{2-14} 
                                                                                             
                                                                                             & \begin{tabular}[c]{@{}l@{}}Cumulative Attributes\\ Chen \etal \cite{chen2013cumulative}\end{tabular}   & 2.07                 &   6.86                  &  3.43                   &     17.07                &                 &                &                                               &                               &                                        &                                       &                                        &                                       \\ \cline{2-14} 
                                                                                              & \begin{tabular}[c]{@{}l@{}}Density learning\\ Lempitsky \etal \cite{lempitsky2010learning}\end{tabular}   & 1.7                 &                     &                     &                     & 493.4                & 487.1                &                                               &                               &                                        &                                       &                                        &                                       \\ \cline{2-14} 
                                                                                              & \begin{tabular}[c]{@{}l@{}}Count forest\\ Pham \etal \cite{pham2015count}\end{tabular}                    & 1.61                & 4.40                & 2.5                 & 10.0                &                      &                      &                                               &                               &                                        &                                       &                                        &                                       \\ \cline{2-14} 
                                                                                              & \begin{tabular}[c]{@{}l@{}}Exemplar density\\ Wang \etal \cite{wang2016fast}\end{tabular}                 & 1.98                & 1.82                & 2.74                & {\ul \textbf{2.10}} &                      &                      &                                               &                               &                                        &                                       &                                        &                                       \\ \cline{2-14} 
                                                                                              & \begin{tabular}[c]{@{}l@{}}Random projection forest\\ Xu \etal \cite{xu2016crowd}\end{tabular}            & 1.90                & 6.01                & 3.22                & 15.5                &                      &                      &                                               &                               &                                        &                                       &                                        &                                       \\ \hline
\multirow{7}{*}{\rotatebox[origin=c]{90}{\parbox[c]{9.5cm}{\centering CNN-based approaches}}}   & \begin{tabular}[c]{@{}l@{}}Cross-scene\\ Zhang \etal \cite{zhang2015cross}\end{tabular}                   & 1.60                & 3.31                &                     &                     & 467.0                & 498.5                & 12.9                                          &                               & 181.8                                  & 277.7                                 & 32.0                                   & 49.8                                  \\ \cline{2-14} 
                                                                                              & \begin{tabular}[c]{@{}l@{}}Deep + shallow\\ Boominathan \etal \cite{boominathan2016crowdnet}\end{tabular} &                     &                     &                     &                     & 452.5                &                      &                                               &                               &                                        &                                       &                                        &                                       \\ \cline{2-14} 
                                                                                              & \begin{tabular}[c]{@{}l@{}}M-CNN\\ Zhang \etal \cite{zhang2016single}\end{tabular}                        & {\ul \textbf{1.07}} & {\ul \textbf{1.35}} &                     &                     & 377.6                & 509.1                & 11.6                           &                               & 110.2                  & 173.2                  & 26.4                  & 41.3                  \\ \cline{2-14} 
                                                                                              & \begin{tabular}[c]{@{}l@{}}CNN-boosting\\ Walach and Wolf \cite{walach2016learning}\end{tabular}          & 1.10                &                     & {\ul \textbf{2.01}} &                     & 364.4                &                      &                                               &                               &                                        &                                       &                                        &                                       \\ \cline{2-14} 
                                                                                              & \begin{tabular}[c]{@{}l@{}}Hydra-CNN\\ Onoro \etal \cite{onoro2016towards}\end{tabular}                   &                     &                     &                     &                     & 333.7                & 425.2 &                                               &                               &                                        &                                       &                                        &                                       \\ \cline{2-14} 
                                                                                              & \begin{tabular}[c]{@{}l@{}}Joint local \& global count\\ Shang \etal \cite{skaug2016end}\end{tabular}     &                     &                     &                     &                     & {\ul \textbf{270.3}} &                      & 11.7                                          &                               &                                        &                                       &                                        &                                       \\ \cline{2-14}

                                                                                             & \begin{tabular}[c]{@{}l@{}}MoCNN \\ Kumagai \etal \cite{kumagai2017mixture}\end{tabular}     &                     &                     &     2.75                &                    13.4 & 361.7 & 493.3                      &                                           &                               &                                        &                                       &                                        &  
                                                                                                                              \\ \cline{2-14} 
                                                                                                                              
                                                                                                                                                                                                                           & \begin{tabular}[c]{@{}l@{}}FCN \\ Marsden \etal \cite{marsden2016fully}\end{tabular}     &                     &                     &                     &                     & 338.6 & 424.5                      &                                           &                               & 126.5                                        &       173.5                                &       23.76                                 &  33.12
                                                                                                                              \\ \cline{2-14} 
                                                                                                                              
                                                                                                                              & \begin{tabular}[c]{@{}l@{}}CNN-pixel \\ Kang \etal \cite{kang2017beyond}\end{tabular}     &         1.12            &        2.06             &                     &                     & 406.2 & 404.0                      &                                          13.4 &                               &                                         &                                  &        & 
                                                                                                                              \\ \cline{2-14} 
                                                                                                                              
                                                                                                                              & \begin{tabular}[c]{@{}l@{}}Weighted V-LAD\\ Sheng \etal \cite{sheng2016crowd}\end{tabular}     & 2.86                    &    13.0                &      2.41               &                     9.12 &  &   &                                            &                               &                                       &                                   &       & 
                                                                                                                              \\ \cline{2-14}
                                                                                                                              
                                                                                                                                                                                                                                                            & \begin{tabular}[c]{@{}l@{}}Cascaded-MTL\\ Sindagi \etal \cite{sindagi2017cnnbased}\end{tabular}     &                     &                    &                     &                     & 322.8 & {\ul  \textbf{341.4                    }}  &                                            &                               &    101.3                                     &                                 152.4 &      {\ul  \textbf{20.0}}  & {\ul  \textbf{31.1}}
                                                                                                                              \\ \cline{2-14}

                                                                                              & \begin{tabular}[c]{@{}l@{}}Switching-CNN\\ Sam \etal \cite{sam2017switching}\end{tabular}                & 1.62 & 2.10 &  &  &                     318.1 &        439.2              &                            {\ul  \textbf{9.4}}                   &                               &                         {\ul  \textbf{90.4}   }          &                                     {\ul  \textbf{ 135.0}} &   21.6                                     &                                   33.4 \\ \hline

\end{tabular}%
}
\end{table*}

\noindent \textbf{Shanghai Tech dataset}: Zhang \etal \cite{zhang2016single} introduced a new large-scale crowd counting dataset consisting of 1198 images with 330,165 annotated heads. The dataset is among the largest ones in terms of the number of annotated people and it contains two parts: Part A and Part B. Part A consists of 482 images that are randomly chosen from the Internet whereas Part B consists of images taken from the streets of metropolitan areas in Shanghai. Part A has considerably larger density images as compared to Part B. Both the parts are further divided into training and evaluation sets. The training and test of Part A has 300 and 182 images, respectively, whereas that of Part B has 400 and 316 images, respectively. The dataset successfully attempts to create a challenging dataset with diverse scene types and varying density levels. However, the number of images for various density levels are not uniform making the training and evaluation biased towards low density levels. Nevertheless, the complexities present in this dataset such as varying scales and perspective distortion has created new opportunities for more complex CNN network designs. 

Sample images from the five datasets are shown in Fig. \ref{fig:dataset}. The datasets are also summarized in Table \ref{tab:datasetsummary}. It can be observed that the UCSD and the Mall dataset have relatively low density images and typically focus on single scene type. In contrast, the other datasets have significant variations in the density levels along with different perspectives across images. 

\subsection{Discussion on results}
Results of the recent traditional approaches along with CNN-based methods are tabulated in Table \ref{tab:results}. The count estimation errors are reported directly from the respective original works. The following standard metrics are used to compare different methods:
\begin{equation}
MAE = \frac{1}{N}\sum_{i=1}^{N}|y_i-y'_i|,
\end{equation}
\begin{equation}
MSE = \sqrt{\frac{1}{N}\sum_{i=1}^{N}|y_i-y'_i|^2},
\end{equation}
where MAE is mean absolute error, MSE is mean squared error, $N$ is the number of test samples, $y_i$ is the ground truth count and $y'_i$ is the estimated count corresponding to the $i^{th}$ sample. We make the following observations regarding the results:

\begin{itemize}[noitemsep]
\item In general, CNN-based methods outperform the traditional approaches across all datasets. 
\item While the CNN-based methods are especially effective in large density crowds with a diverse scene conditions, the traditional approaches suffer from high error rates in such scenarios. 
\item Among the CNN-based methods, most performance improvement is achieved by scale-aware and context-aware models. It can be observed from Table \ref{tab:results} that a  reduction in count error is largely driven by the increase in the complexity of CNN models (due to addition of context and scale information). 
\item While the multi-column CNN architecture \cite{zhang2016single} achieves the state-of-the-art results on 3 datasets: UCSD, WorldExpo '10 and ShanghaiTech, the CNN-boosting approach by \cite{walach2016learning} achieves the best results on the Mall dataset. The best results on the UCF\textunderscore CC\textunderscore 50 dataset are achieved by joint local and global count approach \cite{skaug2016end} and Hydra-CNN \cite{onoro2016towards}. 
\item The work in \cite{walach2016learning} suggests that layered boosting can achieve performances that are comparable to scale aware models. 
\item The improvements obtained by selective sampling in \cite{wang2015deep} and \cite{walach2016learning} suggests that it helps to obtain unbiased performance. 
\item Whole image-based methods such as Zhang \etal \cite{zhang2016single} and Shang \etal \cite{skaug2016end} are less computationally complex from the prediction point of view and they have proved to achieve better results over patch-based techniques. 
\item Finally, techniques such as layered boosting and selective sampling \cite{onoro2016towards,wang2016fast} not only improve the estimation error but also reduce the training time significantly. 
\end{itemize}

\begin{figure}[htp!]
\begin{center}
\begin{minipage}{0.33\linewidth}
\includegraphics[width=\linewidth]{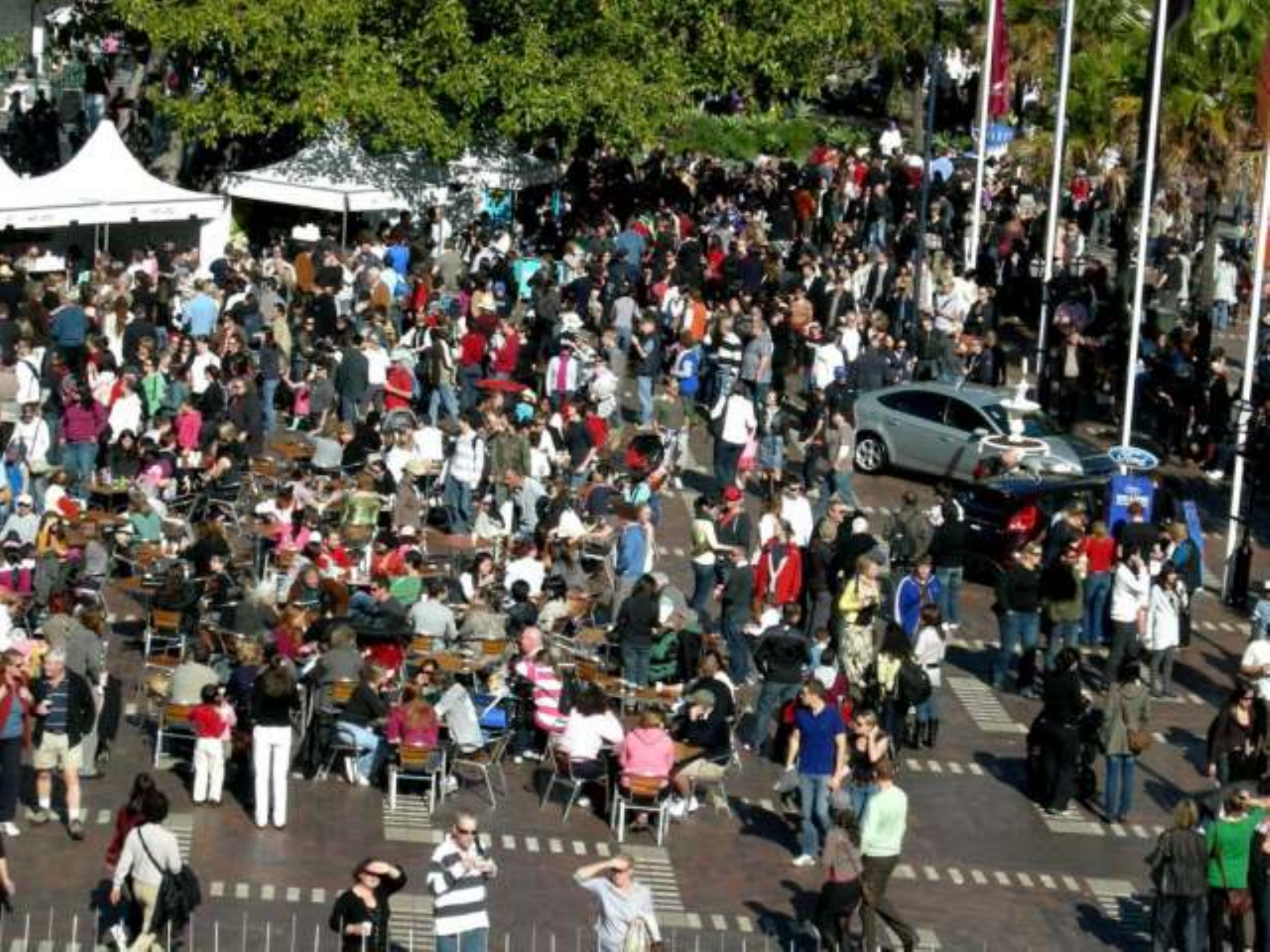}
\end{minipage}%
\hfill
\begin{minipage}{0.33\linewidth}
\includegraphics[width=\linewidth]{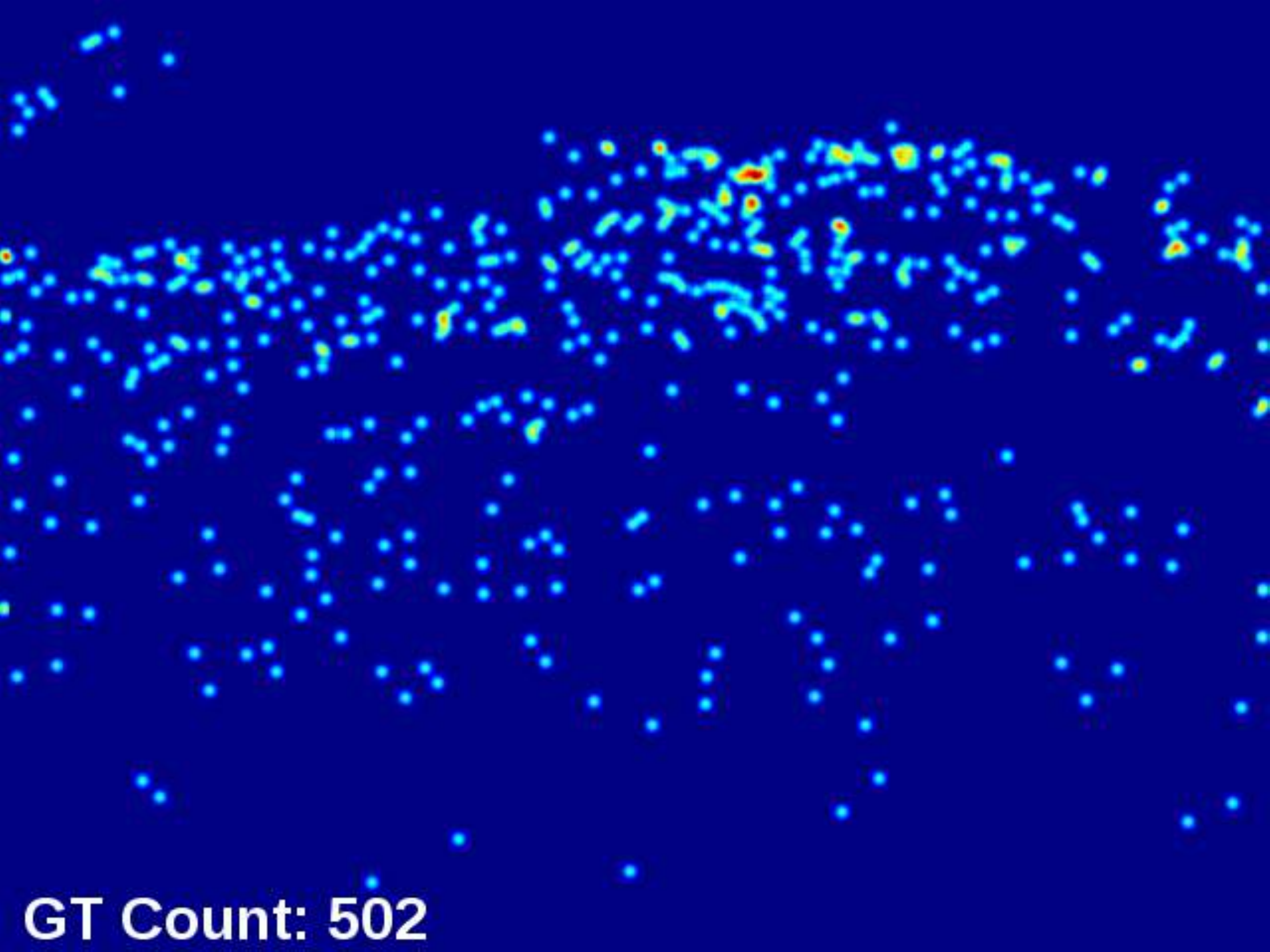}
\end{minipage}
\begin{minipage}{0.33\linewidth}
\includegraphics[width=\linewidth]{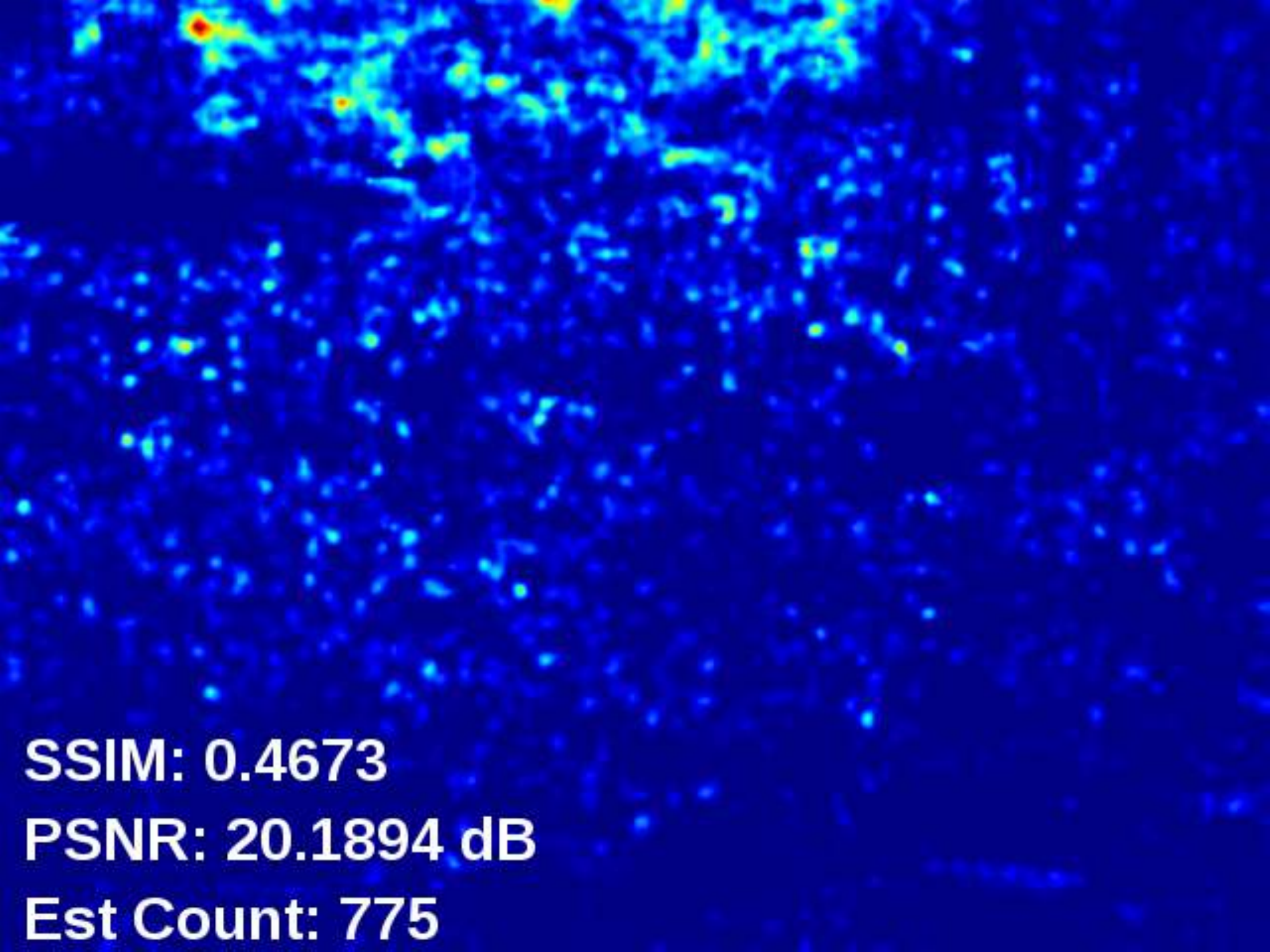}
\end{minipage}
\end{center}

\begin{center}
\begin{minipage}{0.33\linewidth}
\includegraphics[width=\linewidth]{Fig27}
\captionof*{figure}{(a)}
\end{minipage}%
\hfill
\begin{minipage}{0.33\linewidth}
\includegraphics[width=\linewidth]{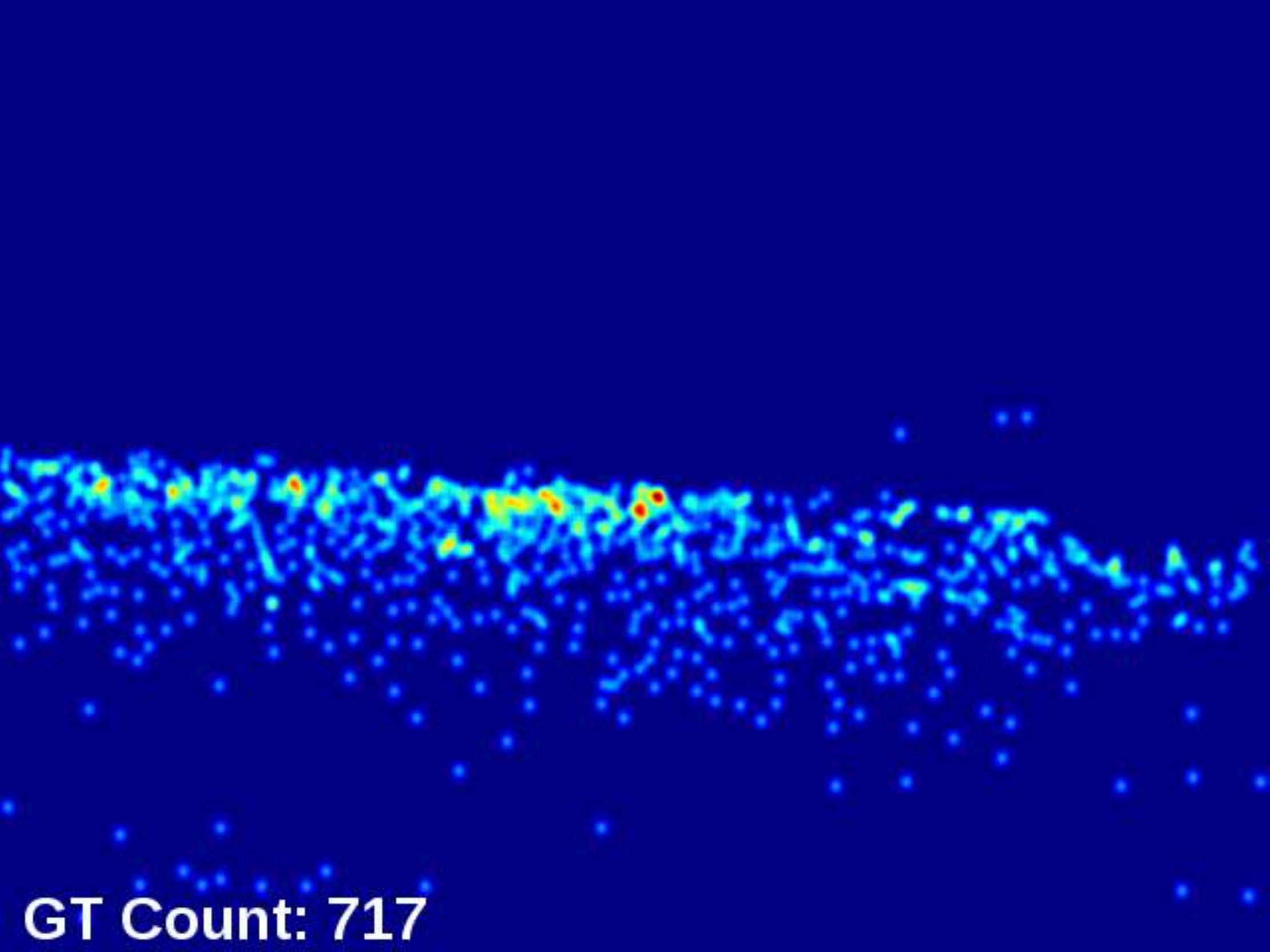}
\captionof*{figure}{(b)}
\end{minipage}
\begin{minipage}{0.33\linewidth}
\includegraphics[width=\linewidth]{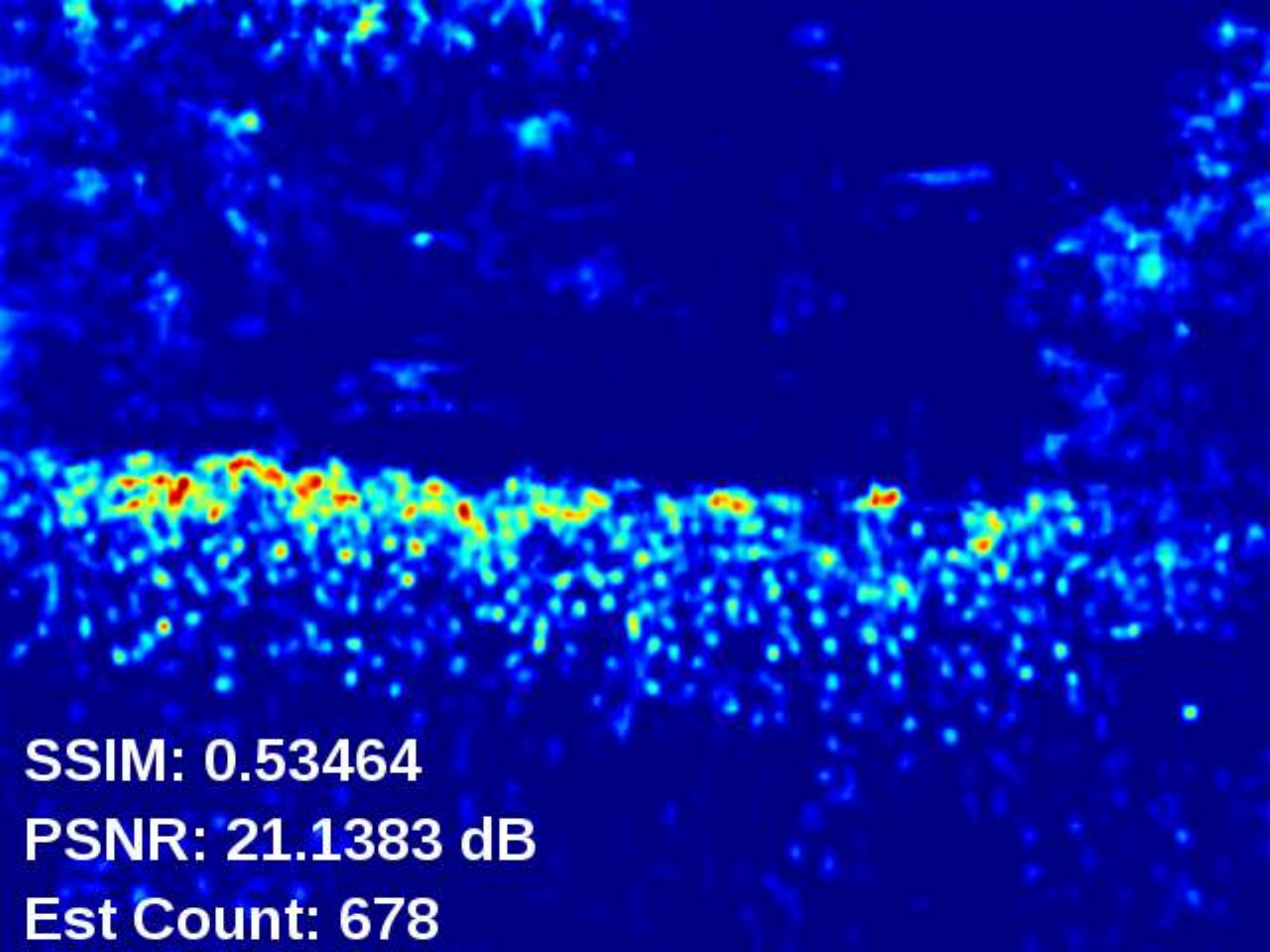}
\captionof*{figure}{(c)}
\end{minipage}
\end{center}
\vskip-10pt \captionof{figure}{Results of Zhang \etal \cite{zhang2016single} on ShanghaiTech dataset. (a) Input image(b) Ground-truth density map (c) Estimated density maps. It can be observed that though the method is able to accurate estimation of crowd count, the estimated density maps are of poor quality.}
\label{fig:quality}
\end{figure}

\begin{figure}[htp!]
\begin{center}
\begin{minipage}{1\linewidth}
\includegraphics[width=0.47\linewidth]{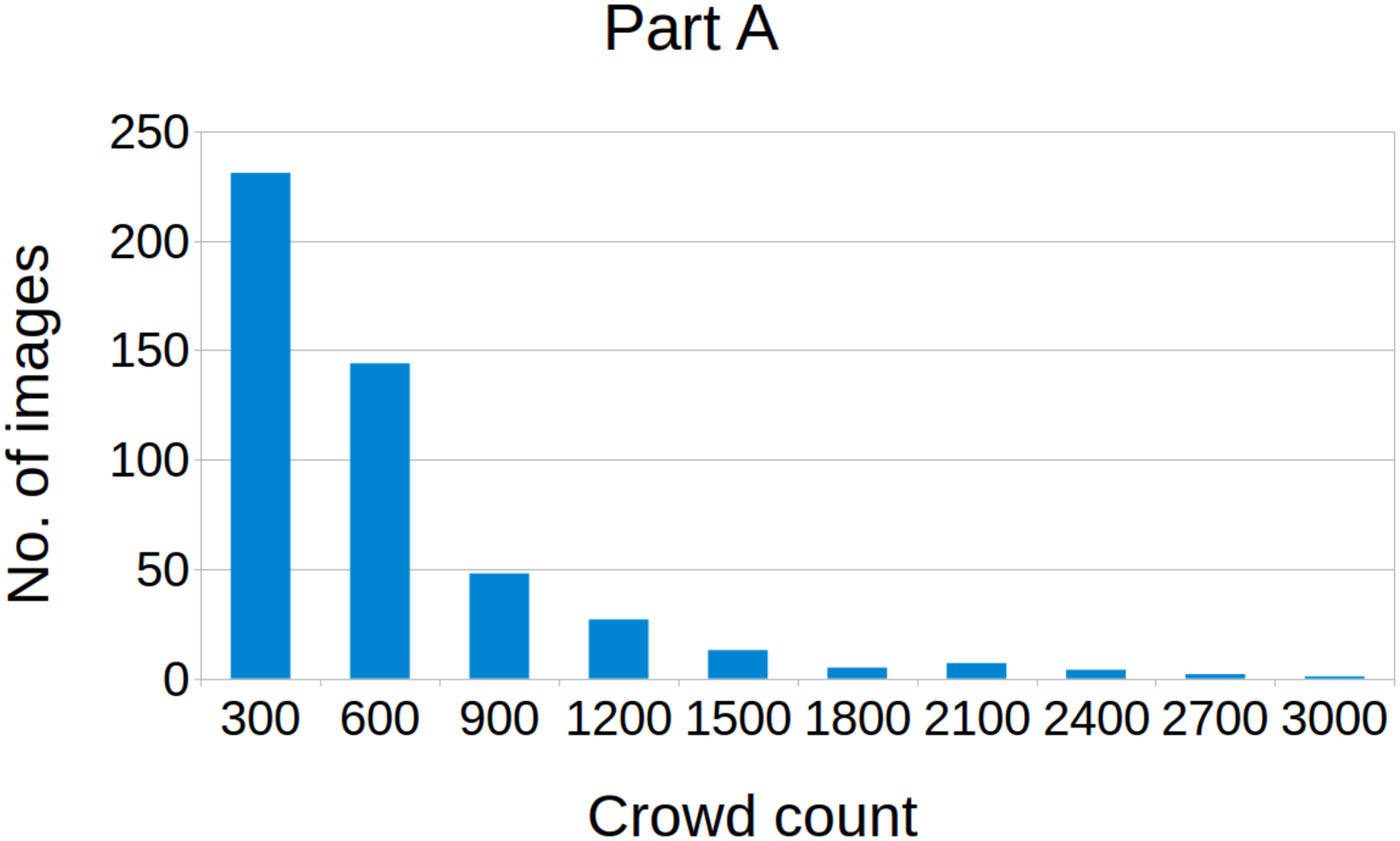}
\includegraphics[width=0.47\linewidth]{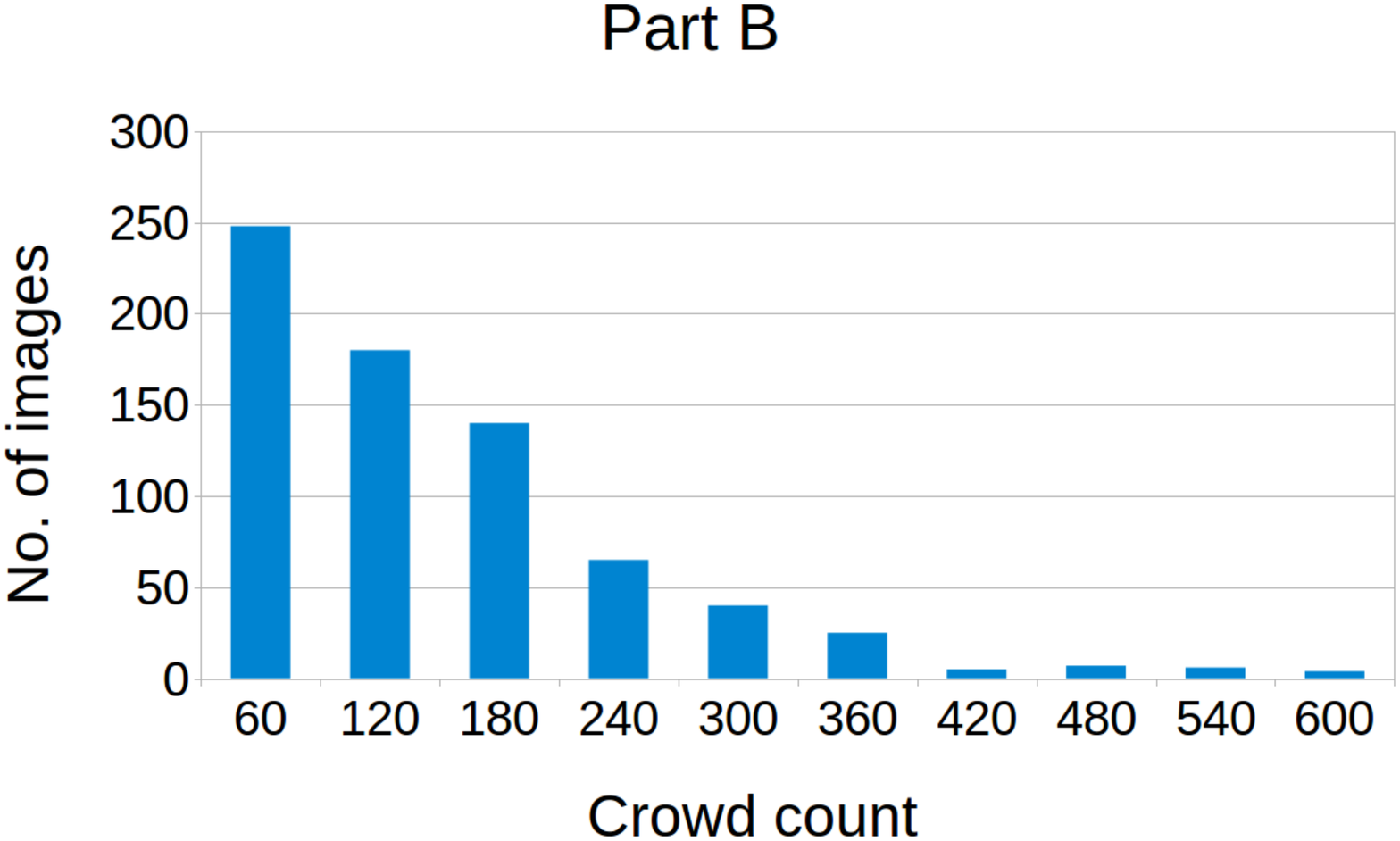}
\end{minipage}%
\vskip-8pt
\captionof{figure}{Distribution of crowd counts in ShanghaiTech dataset. It can be observed that the dataset is highly imbalanced. }
\label{fig:shanghaitech}
\end{center}
\end{figure}

\section{Future research directions}
\label{sec:future_research}

Based on the analysis of various methods and results from Section \ref{sec:survey_cnn} and \ref{sec:datasets_and_results} and the trend of other developments in computer vision, we believe that CNN-based deeper architectures will dominate further research in the field of crowd counting and density estimation. We make the following observations regarding future trends in research on crowd counting:

\begin{enumerate}[noitemsep]
\item Given the requirement of large datasets for training deep networks, collection of large scale datasets (especially for extremely dense crowds) is essential. Though many datasets exist currently, only one of them (The UCF\textunderscore CC\textunderscore 50 \cite{idrees2013multi}) caters to large density crowds. However, the size of the dataset is too small for training deeper networks. Though Shanghai Tech \cite{zhang2016single}) attempts to capture large density crowds, the number of images per density level is non-uniform with a large number of images available for low density levels and very few samples for high density levels (as shown in Fig. \ref{fig:shanghaitech}). 

\item Considering the difficulty of training deep networks for new scenes, it would be important to explore how to leverage from models trained on existing sources. Most of the existing methods retrain their models on a new scene and it is impractical to do so in real world scenarios as it would be expensive to obtain annotations for every new scene. Zhang \etal \cite{zhang2015cross} attempted to address this issue by performing a data driven training without the need of labelled data for new scenes. In an another approach, Liu \etal \cite{liu2015bayesian} considered the problem of transfer learning for crowd counting. A model adaptation technique for Gaussian process counting model was introduced. Considering the source model as a prior and the target dataset as a set of observations, the components are combined into a predictive distribution that captures information in both the source and target datasets. However, the idea of transfer learning or domain adaptation \cite{VMP_SPM_DA_2015} for crowd scenes is relatively unexplored and is a nascent area of research.

\item Most crowd counting and density estimation methods have been designed for and evaluated either only on single images or videos.  Combining the techniques developed separately for these methods is a non-trivial task. Development of low-latency methods that can operate in real-time for counting people in crowds from videos is another interesting problem to be addressed in future.

\item Another key issue ignored by earlier research is that the quality of estimated crowd density maps. Many existing CNN-based approaches have a number of max-pooling layers in their networks compelling them to regress on down-sampled density maps. Also, most methods optimize over traditional Euclidean loss which is known to have certain disadvantages \cite{johnson2016perceptual1}. Regressing on down-sampled density maps using Euclidean loss results in low quality density maps. Fig. \ref{fig:quality} demonstrates the results obtained using the state-of-the-art method \cite{zhang2016single}. It can be observed that though accurate count estimates are obtained, the quality of the density maps is poor. As a result, these poor quality maps adversely affect other higher level cognition tasks which depend on them. Recent work on style-transfer \cite{zhang17ICCV}, image de-raining \cite{zhang2017image} and image-to-image translation \cite{pix2pix2016} have demonstrated promising results from the use of additional loss functions such as adversarial loss and perceptual loss. In principle, density estimation can be considered as an image-to-image translation problem and it would be interesting to see the effect of these recent loss functions.  Generating high quality density maps along with low count estimation error would be another important issue to be addressed in the future.

\item Finally, considering advancements by scale-aware \cite{zhang2016single,onoro2016towards}  and context-aware models \cite{skaug2016end}, we believe designing networks to incorporate additional contextual and scale information will enable further progress.

\end{enumerate}

\section{Conclusion}
\label{sec:conclusion}
This article presented an overview of recent advances in CNN-based methods for crowd counting and density estimation.   In particular, we summarized various methods for crowd counting into traditional approaches (that use hand-crafted features) and CNN-based approaches. The CNN-based approaches are further categorized based on the training process and the network property. Obviously all the literature on crowd counting cannot be covered, hence, we have chosen a representative subset of the latest approaches for a detailed analysis and review. We also reviewed the results demonstrated by various traditional and CNN-based approaches to conclude that CNN-based methods are more adept at handling large density crowds with variations in object scales and scene perspective. Additionally, we observed that incorporating scale and contextual information in the CNN-based methods drastically improves the estimation error. Finally, we identified some of the most compelling challenges and issues that confront research in crowd counting and density estimation using computer vision and machine leaning approaches.  

\vspace{5pt}
\noindent \textbf{Acknowledgement}
This work was supported by US Office of Naval Research (ONR) Grant YIP N00014-16-1-3134.

\bibliographystyle{model2-names}
\bibliography{refs}
\end{document}